\documentclass[11pt]{article}
\usepackage[margin=1in]{geometry}

\usepackage{amsmath,amssymb}
\usepackage{graphicx}
\usepackage{caption}
\usepackage{tikz-cd}
\usepackage{multicol}
\usepackage{tabularx}
\usepackage{booktabs}
\usepackage{longtable}
\usepackage{pdflscape}
\usepackage{url}
\usepackage[most]{tcolorbox}
\usepackage{listings}

\lstdefinestyle{promptstyle}{
    basicstyle=\ttfamily\footnotesize,
    breaklines=true,
    columns=fullflexible,
    keepspaces=true,
    showstringspaces=false
}

\newtcblisting{promptbox}{
    enhanced,
    breakable,
    listing only,
    colback=black!2,
    colframe=black!45,
    boxrule=0.5pt,
    arc=2pt,
    left=6pt,
    right=6pt,
    top=6pt,
    bottom=6pt,
    listing options={style=promptstyle}
}

\title{Learning Scientific Trajectories to Predict Future Biomedical Discoveries} 
\author{%
Chia-Hsuan Chang$^{1,\dagger}$, Brian Ondov$^{1,\dagger}$, Jiang Lv$^{2}$, Lan Jiang$^{1}$, Yunmai Wang$^{1}$,\\
Haerin Song$^{3}$, Xueqing Peng$^{1}$, Qiaozhu Mei$^{4}$, Xiaolei Su$^{2,*}$, Hua Xu$^{1,*}$\\[0.75em]
\parbox{0.92\textwidth}{\centering\small
$^{1}$Department of Biomedical Informatics and Data Science, Yale School of Medicine, New Haven, CT 06510, USA\\
$^{2}$Department of Cell Biology, Yale School of Medicine, New Haven, CT 06520, USA\\
$^{3}$Interdisciplinary Program in Artificial Intelligence, Seoul National University, 1 Gwanak-ro, Gwanak-gu, Seoul 08826, Republic of Korea\\
$^{4}$School of Information, University of Michigan, Ann Arbor, MI 48109, USA\\[0.5em]
$^{\dagger}$These authors contributed equally.\\
$^{*}$Co-corresponding authors.}
}
\date{\today}

\begin{document}
\maketitle

\begin{abstract}
AI co-scientists have expanded the use of large language models for generating scientific hypotheses and guiding experimental investigation. Here we investigate whether temporal patterns in the semantic representations of complete biomedical conclusions can guide hypothesis generation. We introduce ForeSCITE, a framework that identifies thematic clusters of published conclusions, estimates a direction separating earlier from more recent findings within each cluster, and projects existing conclusion embeddings along these directions. An embedding language model trained to reconstruct conclusions decodes the projected coordinates into natural-language hypotheses. Across 407 clusters derived from approximately three million pre-cutoff conclusions, we identify temporal structure and illustrate its retention after decoding. In retrospective, time-sliced evaluations, ForeSCITE increases semantic coverage beyond the historical corpus, biomedical entity-pair novelty and post-cutoff pair re-observation relative to five prompt-based and retrieval-augmented baselines. Expert review of prioritized CAR-T hypotheses identified 37 candidates as both novel and biologically plausible, and expert justifications for five selected hypotheses illustrate their biological rationale and subsequent literature support where available. These findings support temporal guidance in semantic space as a scalable mechanism for proposing biomedical hypotheses for subsequent scientific evaluation.
\end{abstract}

\section{Introduction}

AI co-scientists are beginning to connect literature synthesis and hypothesis generation with experimental investigation. Systems built on large language models (LLMs) have proposed therapeutic candidates, investigated biological mechanisms and designed biomolecules whose properties were subsequently examined experimentally~\cite{gottweisAcceleratingScientificDiscovery2026,ghareebMultiagentSystemAutomating2026,swansonVirtualLabAI2025}. These advances extend the role of AI from analyzing scientific evidence to helping determine what should be investigated next. As biomedical knowledge expands across specialized research areas~\cite{gonzalez-marquezLandscapeBiomedicalResearch2024}, identifying promising directions within the space of possible hypotheses becomes an increasingly consequential computational problem. The chronological development of the literature offers a potential source of guidance: changes in what researchers report may indicate which extensions of existing findings warrant exploration.

Many recent hypothesis-generation systems combine literature retrieval with LLM generation and iterative critique~\cite{baekResearchAgentIterativeResearch2025,wangSciMONScientificInspiration2024}. These workflows support flexible reasoning within a specified research context, with exploration guided by research objectives, retrieved evidence and successive revisions. Alongside its role as evidence for this reasoning, the literature records how scientific claims change over time. Chronological differences among findings may reveal emerging mechanisms, interventions and experimental settings that are difficult to discern from individual publications. We ask whether these changes can provide an explicit direction for hypothesis generation: can temporal patterns in the semantic representations of complete biomedical conclusions guide the generation of detailed, testable claims?

Literature-based discovery (LBD) provides a foundation for this question by uncovering connections across otherwise separate bodies of biomedical evidence~\cite{swansonFishOilRaynauds1986,henryLiteratureBasedDiscovery2017}. Temporal LBD methods have further demonstrated the value of historical change for predicting future connections between biomedical concepts~\cite{zhouLearningTemporalDifference2022,zhouGeneratingBiomedicalHypothesis2024,shariatmadariConceptDriftLeveragingSpatial2025}. Related work has forecast the motion of research-topic embeddings and decoded predicted representations into topic phrases~\cite{huangFrameworkDemonstratingForecasting2025}, while materials-science models have used evolving concept graphs and semantic information to predict emerging concept combinations~\cite{marwitzPredictingNewResearch2026}. These approaches establish relevant precedents for temporal prediction in scientific knowledge representations. Our focus is the use of temporal directions in representations of complete biomedical conclusions to generate natural-language hypotheses. Such hypotheses can express an intervention, biological context, direction of effect and proposed consequence within a single claim, providing a richer object for subsequent scientific evaluation than a topic phrase or concept pair alone.

\begin{figure}[!htbp]
    \centering
    \includegraphics[width=\textwidth]{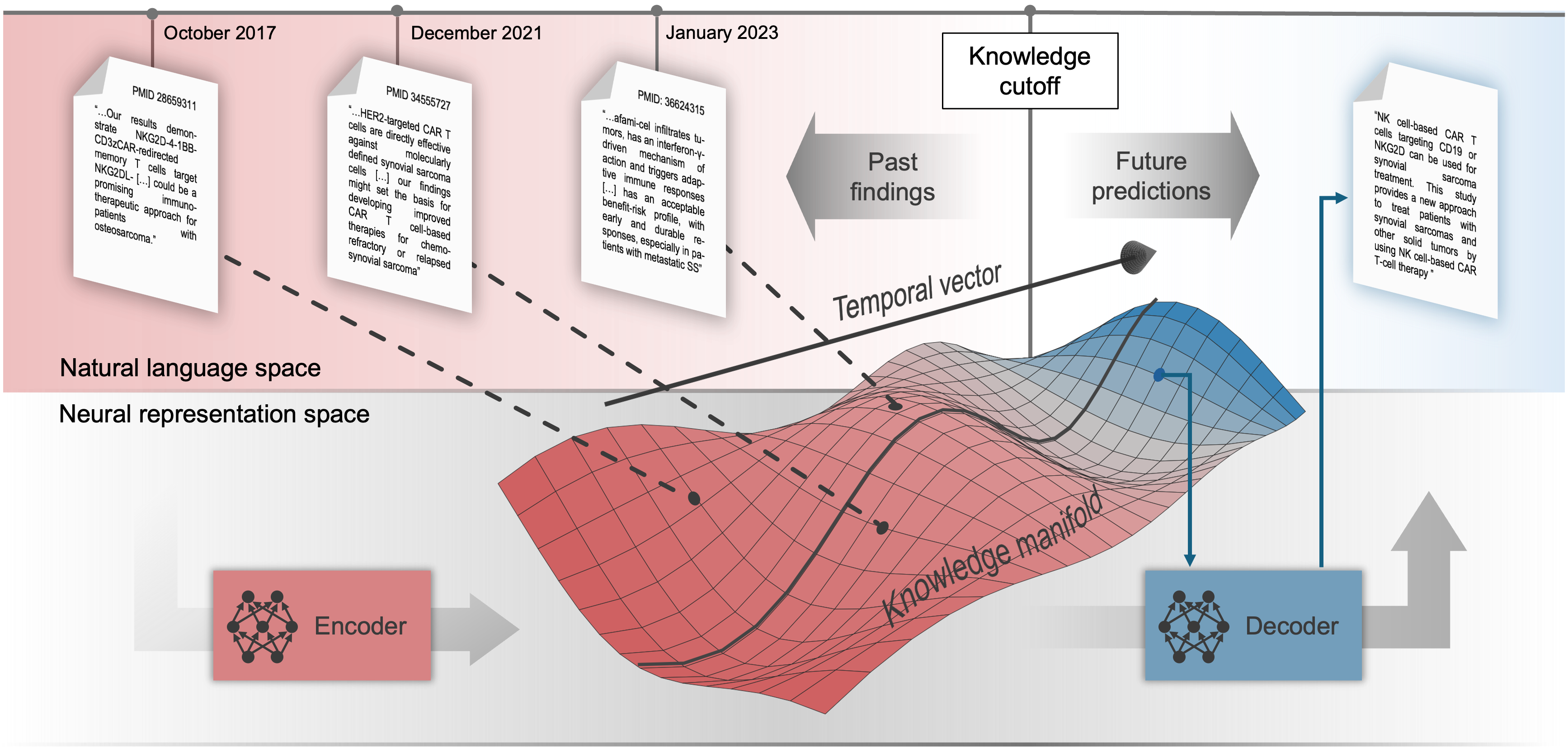}
    \caption{\textbf{Temporal extrapolation of biomedical findings for hypothesis generation.} Published conclusions in natural-language space are encoded as points in a shared semantic representation space. A temporal direction learned from historical findings guides extrapolation from a reference conclusion toward a candidate representation, which an embedding language model decodes into a natural-language hypothesis. Red and blue regions distinguish historical findings and candidate future findings, respectively, separated by the knowledge cutoff. The CAR-T example illustrates how prior findings can inform a new disease-specific hypothesis. The surface and temporal arrow are conceptual illustrations of the embedding space and learned extrapolation direction.}
    \label{fig:forescite_concept}
\end{figure}

Here we introduce ForeSCITE (\textbf{Fore}casting \textbf{Sci}ence with \textbf{T}emporal \textbf{E}xtrapolation), a framework for biomedical hypothesis generation through temporal extrapolation in a semantic space of published conclusions (Fig.~\ref{fig:forescite_concept}). ForeSCITE embeds historical conclusions and organizes them into thematic clusters. Within each cluster, a linear classifier distinguishes earlier from more recent conclusions to estimate a research trend direction in the original embedding space. The framework projects an existing conclusion embedding along this direction and uses an embedding language model (ELM), trained to reconstruct conclusions from their embeddings, to decode the projected coordinate into a candidate hypothesis~\cite{tennenholtzDemystifyingEmbeddingSpaces2024,ondov-etal-2026-coreelm}. This formulation uses the temporal organization of published claims to supply a learned direction for generation, with an adjustable step size controlling the extent of extrapolation. It enables systematic exploration from historical findings while retaining the expressive scope of natural-language hypotheses.

Across 407 thematic clusters derived from approximately three million pre-cutoff biomedical conclusions, we identify temporal structure and illustrate its retention through the decoding and re-embedding of projected representations. In retrospective, time-sliced evaluations, ForeSCITE expands semantic coverage beyond the historical corpus and increases biomedical entity-pair novelty relative to the evaluated prompt-based and retrieval-augmented baselines. It also increases the fraction of generated pairs absent from the historical corpus but observed in post-cutoff literature. Expert assessment of CAR-T hypotheses identified 37 candidates as both novel and biologically plausible, and justifications for five selected hypotheses illustrate how generated ideas connect to biological mechanisms and subsequent literature where available. These findings support temporal guidance in conclusion-level semantic space as a mechanism for proposing biomedical hypotheses. ForeSCITE could complement AI co-scientists by supplying candidates for literature-based scrutiny, prioritization and experimental development.

\section{Results}

\subsection{ForeSCITE generates hypotheses by extrapolating trajectories of biomedical findings}
\label{sec:forescite_overview}

\begin{figure}[!htbp]
    \centering
    \includegraphics[width=\textwidth,height=0.74\textheight,keepaspectratio]{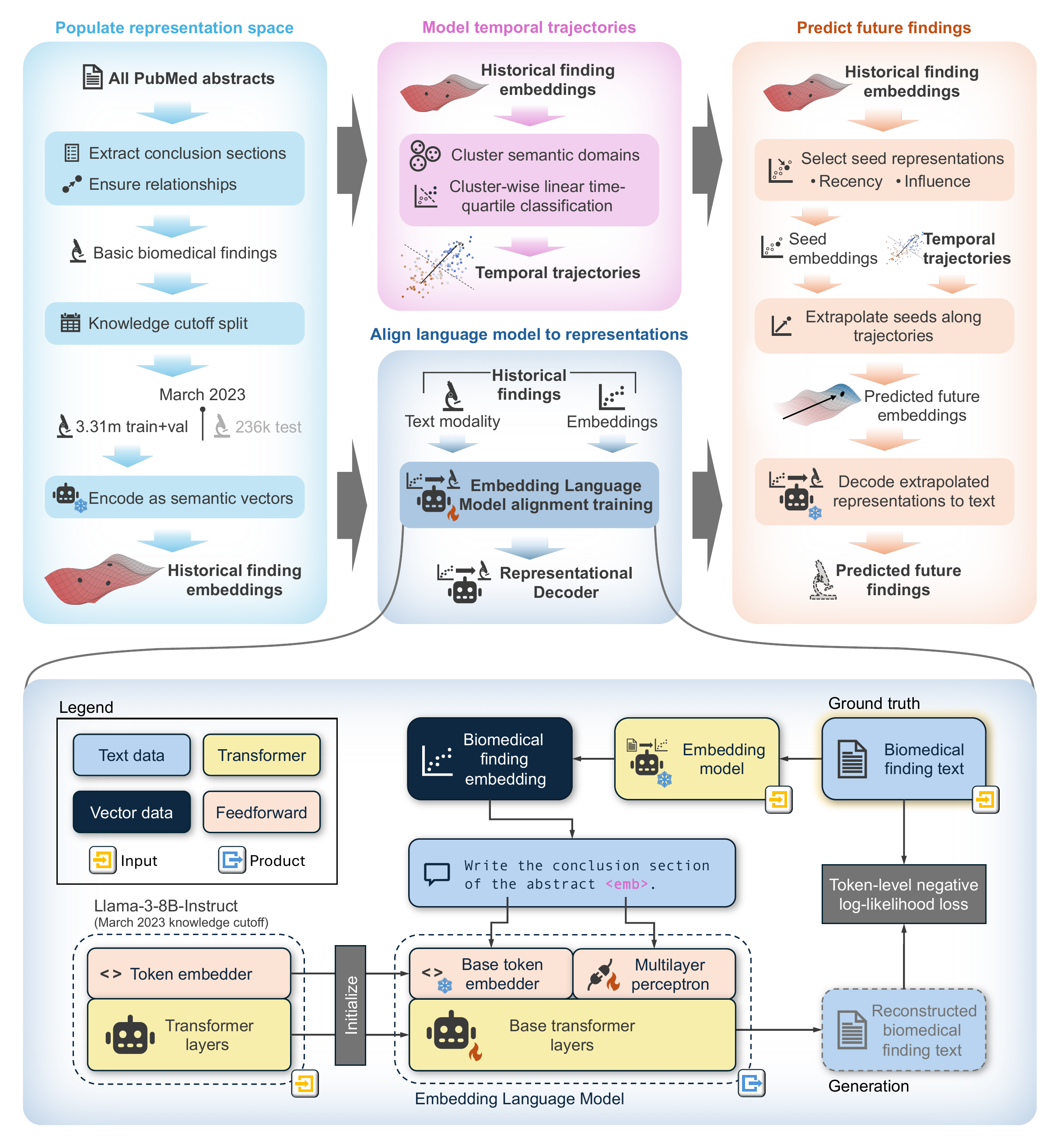}
    \caption{\textbf{ForeSCITE workflow from historical literature to candidate biomedical hypotheses.} \textbf{Left,} Conclusions extracted from PubMed abstracts are filtered for biological relations, partitioned by the knowledge cutoff and encoded as semantic vectors. \textbf{Upper middle,} Thematic clustering and early--late classification identify research trend directions. \textbf{Lower middle,} Alignment of paired conclusion texts and embeddings produces an ELM that serves as a representation-to-text decoder. \textbf{Right,} Reference conclusions selected by recency or citation impact are extrapolated along the corresponding temporal directions and decoded into candidate hypotheses. \textbf{Bottom,} ELM alignment reconstructs the original conclusion text from its embedding through an adapter and language model. Snowflakes and flames denote frozen and trainable components, respectively. FFN, feed-forward network. Implementation and training details are provided in Methods.}
    \label{fig:framework}
\end{figure}

The ForeSCITE workflow combines a literature-derived representation of biomedical findings, learned research trend directions and an ELM that converts representations back into text (Fig.~\ref{fig:framework}). Historical conclusions are organized into thematic domains, providing a domain-specific context for extrapolation. Reference findings selected by recency or citation impact are shifted along the corresponding research trend direction, with an adjustable step size controlling the extent of displacement. The ELM translates these extrapolated representations into readable hypotheses. Details of corpus preparation, trajectory estimation, ELM training and hypothesis generation are provided in Methods.

We evaluated ForeSCITE across 407 thematic clusters derived from approximately three million pre-cutoff conclusions. The following analyses test whether conclusion embeddings contain a temporal signal and retain temporal movement after decoding, whether extrapolation expands semantic and relational coverage toward post-cutoff literature, and whether generated CAR-T hypotheses are judged novel and biologically plausible by domain experts.

\subsection{Conclusion embeddings encode publication time and preserve temporal movement after decoding}
\label{sec:temporal_embedding_visualization}

In this section, we first investigated whether the semantic representations of biomedical conclusions encode a temporal signal using linear regression. We then visualized this signal in a two-dimensional representation space. Finally, we examined whether temporal movement can be decoded into natural-language text that preserves the underlying temporal trend.

To first establish the presence of a temporal signal, we examined three illustrative clusters from the 407 clusters identified from 3,005,395 pre-cutoff conclusions: Molecular Hematologic Malignancy Research, Diabetic Vascular and Renal Complications, and Rare Monogenic Disease Genetics (Fig.~\ref{fig:temporal_structure}a, left, middle and right columns, respectively). For each cluster, we used a linear regression model to predict the publication year of each conclusion purely from its 1,024-dimensional semantic representation. The top row of Fig.~\ref{fig:temporal_structure}a shows the distributions of predicted years grouped by actual publication period, with pre- and post-cutoff conclusions distinguished by color. Across all three clusters, predicted years generally increase with actual publication period, and post-cutoff conclusions tend to have higher predicted years than pre-cutoff conclusions. The overlap between periods and the range of predicted years indicate a graded temporal signal rather than exact date recovery or complete separation between historical and future conclusions.

To quantify temporal separability beyond these three illustrative clusters, we used conclusions from the earliest and latest publication-year quartiles within each of the 407 thematic clusters to fit a linear temporal classifier and measured its accuracy by five-fold cross-validation. The classifiers distinguished early from recent conclusions with high accuracy (mean $=0.847$, s.d. $=0.048$; Fig.~\ref{fig:temporal_structure}b), supporting the estimation of research trend directions within thematic clusters. Accuracy also increased with the log-transformed number of early and late conclusions available for training (slope $=0.0421$, intercept $=0.7043$, $r=0.364$, $P=3.69\times10^{-14}$; Fig.~\ref{fig:temporal_structure}c), indicating an association between greater training-data availability and higher temporal classification accuracy.

This temporal signal suggests that semantic representations do not merely encode information about publication time, but may organize conclusions along a temporal trajectory. We therefore sought to make this trajectory explicit in a two-dimensional representation space. This was achieved by training a multilayer perceptron (MLP) to predict publication year, with a progressively compressed architecture that reduced the 1,024-dimensional semantic representations through 16 and 8 dimensions to a two-dimensional bottleneck. The bottom row of Fig.~\ref{fig:temporal_structure}a shows these bottleneck coordinates, with hexagonal bins colored by publication year. The representations reveal a progression from earlier to later conclusions, with cluster-specific differences in the width and shape of the occupied region.

Having identified the presence of a temporal signal, we then investigated whether temporal movement can be preserved during the conversion of extrapolated vectors into natural-language text. For each displayed cluster, we extrapolated ten sampled recent pre-cutoff conclusion embeddings along its research trend direction with step size $\alpha=0.25$, decoded the projected embeddings to natural language using the trained ELM, and re-embedded the decoded conclusions with the original conclusion encoder. We passed these re-embedded texts through the same trained MLP and placed them in the reference visualization without refitting the mapping. In the bottom row of Fig.~\ref{fig:temporal_structure}a, gray lines connect light-blue reference markers to red markers representing decoded hypotheses. These paired movements generally follow the early-to-late direction, supporting retention of temporal displacement through the vector-to-text-to-vector transformation. We further extrapolated the most recent 5\% of pre-cutoff conclusions within each displayed cluster and included their decoded, re-embedded representations as the separate red ``Extrapolated'' group in the top row. This group generally received higher predicted years than the existing conclusions, providing a complementary distribution-level view of temporal movement after decoding. These predicted years describe the models' temporal positioning of generated text, not actual future publication dates.

Together, these results demonstrate that temporal signal is present in biomedical conclusion embeddings and can be preserved through vector extrapolation, natural-language decoding, and re-embedding. This provides a basis for ForeSCITE to leverage temporal movement in the embedding space to steer generated hypotheses toward future biomedical conclusions.

\begin{figure}[!htbp]
    \centering
    \setlength{\parskip}{0pt}
    \textbf{a}\par\vspace{0.5ex}
    \includegraphics[width=\linewidth]{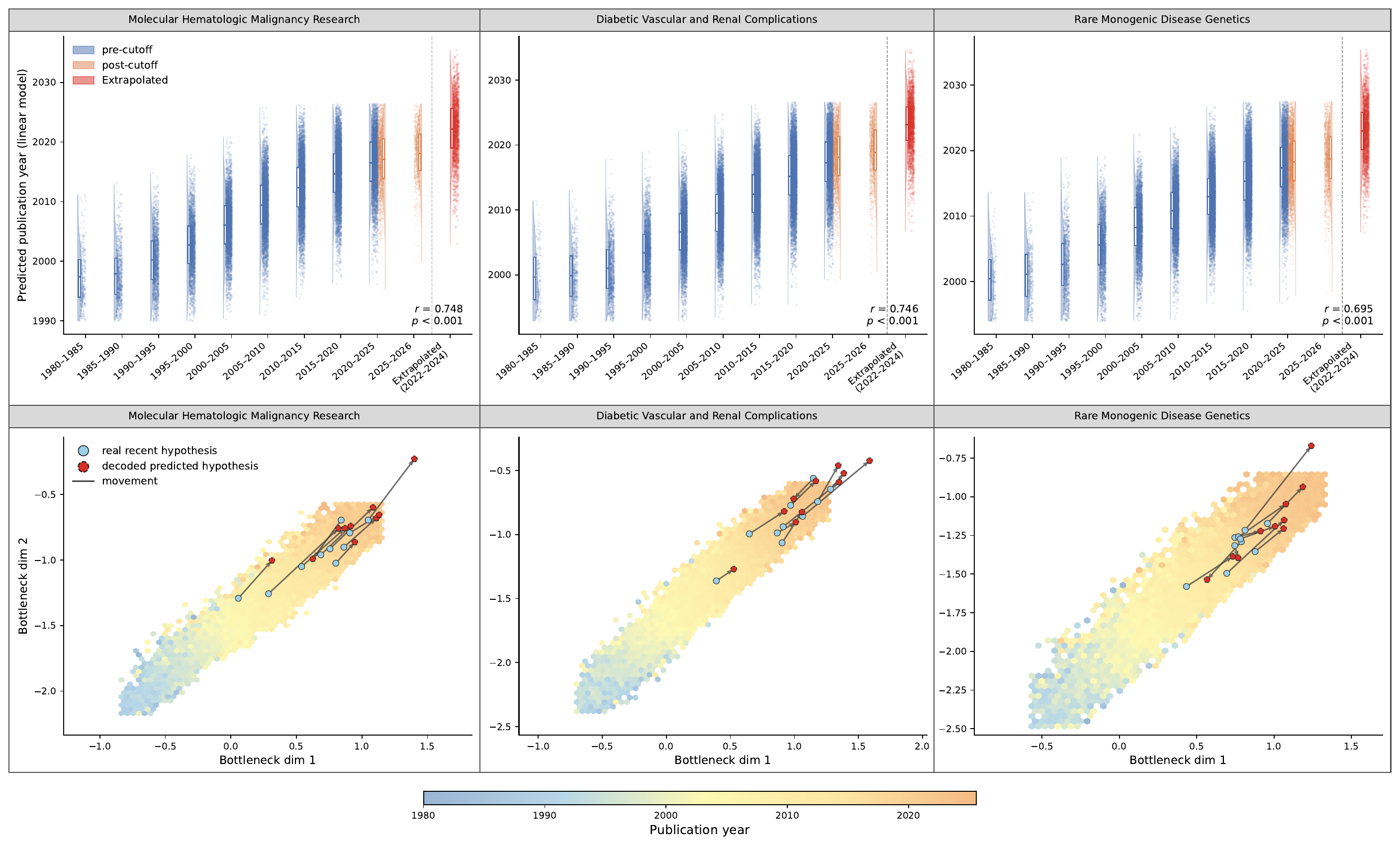}
    \caption[Publication-time structure, decoded temporal movement and early--late classification of biomedical conclusions.]{\textbf{Publication-time structure, decoded temporal movement and early--late classification of biomedical conclusions.} \textbf{a,} Publication-time structure and decoded temporal movement across three illustrative thematic clusters. Columns show Molecular Hematologic Malignancy Research (left), Diabetic Vascular and Renal Complications (middle), and Rare Monogenic Disease Genetics (right). \textbf{Top row,} Distributions of publication years predicted by linear regression on 1,024-dimensional conclusion embeddings, grouped by actual publication period. Blue and orange denote pre- and post-cutoff conclusions, respectively. The separate red ``Extrapolated'' group shows predictions for decoded, re-embedded hypotheses generated from the most recent 5\% of pre-cutoff conclusions in each cluster; the parenthetical years identify the source period, not publication dates of the generated hypotheses. \textbf{Bottom row,} Two-dimensional bottleneck coordinates of a publication-year-predicting MLP, with hexagonal bins colored by publication year according to the shared color bar. Light-blue markers denote ten sampled recent pre-cutoff reference conclusions per cluster, red markers denote re-embedded ELM-decoded hypotheses after extrapolation along the cluster-specific research trend direction with $\alpha=0.25$, and gray lines connect matched pairs. Both endpoints are mapped by the same fitted MLP without refitting. The paired movements generally follow the early-to-late gradient.}
    \label{fig:temporal_structure}
\end{figure}

\begin{figure}[!htbp]
    \ContinuedFloat
    \centering
    \setlength{\parskip}{0pt}
    \begin{minipage}[t]{0.48\textwidth}
        \textbf{b}\par\vspace{0.5ex}
        \includegraphics[width=\linewidth]{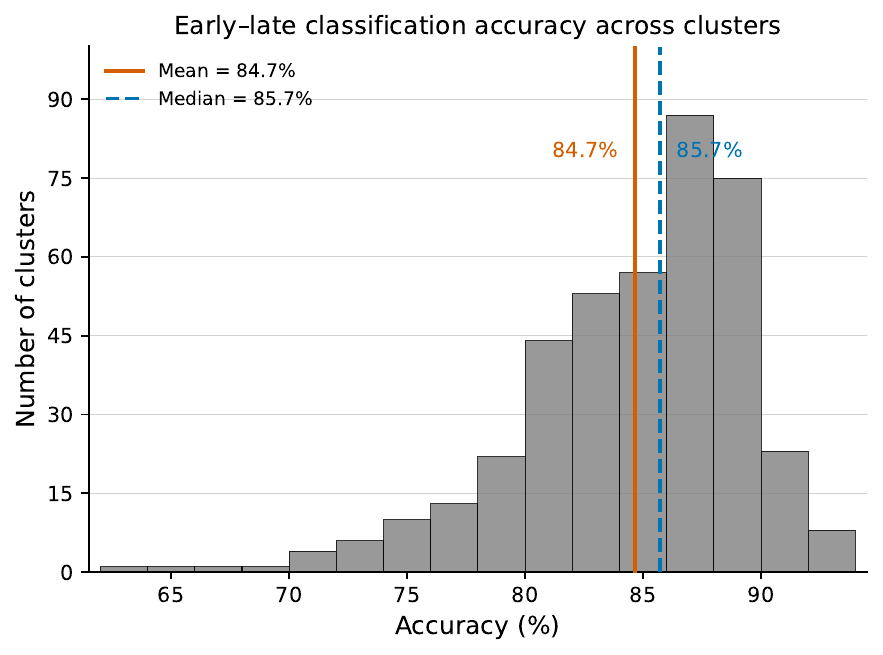}
    \end{minipage}\hfill
    \begin{minipage}[t]{0.48\textwidth}
        \textbf{c}\par\vspace{0.5ex}
        \includegraphics[width=\linewidth]{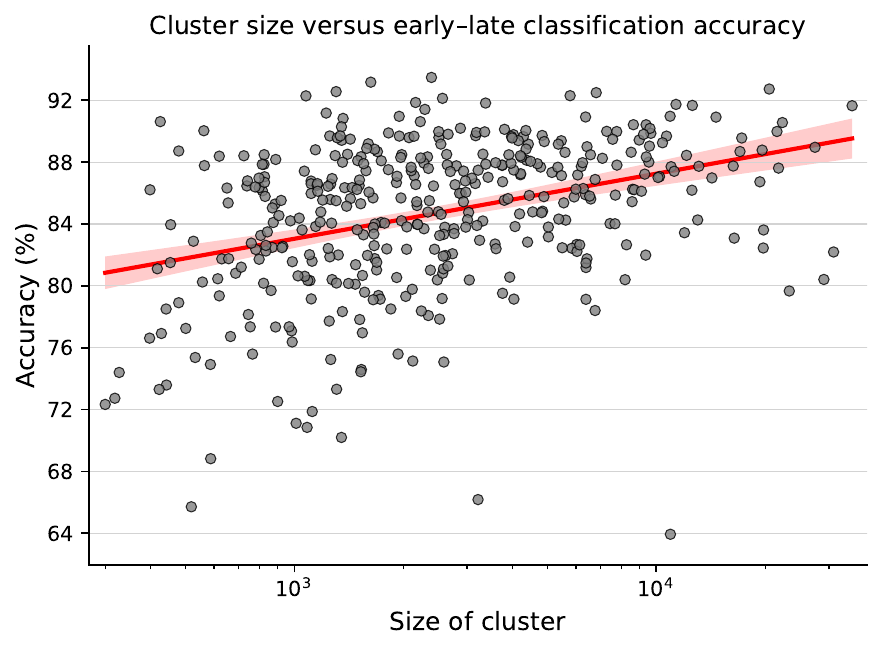}
    \end{minipage}
    \caption[]{\textbf{Publication-time structure, decoded temporal movement and early--late classification of biomedical conclusions (continued).} \textbf{b,} Distribution of five-fold cross-validation accuracy for linear classifiers distinguishing early from recent conclusions across the 407 thematic clusters (mean $=0.847$, s.d. $=0.048$; median $=0.857$). \textbf{c,} Accuracy versus cluster size, defined here as the combined number of early- and late-period conclusions used to train the classifier. Each point denotes one cluster. The red line shows a regression on log-transformed cluster size and the shaded region its 95\% confidence interval (slope $=0.0421$, intercept $=0.7043$, $r=0.364$, $P=3.69\times10^{-14}$).}
\end{figure}

\subsection{Temporal extrapolation expands semantic and relational coverage of future biomedical hypotheses}

We evaluated the effect of ForeSCITE's temporal extrapolation on the semantic and relational coverage of generated hypotheses beyond the pre-cutoff literature using two strategies to select reference conclusions (Methods; Appendix Table~\ref{tab:reference_selection}). In the recency-based analysis, we compared ForeSCITE with three prompt-based approaches (Direct, Temporal, and Novel) and two retrieval-augmented generation approaches (Novel RAG and RAG + Iter.; Methods, Section~\ref{sec:baselines}) using the same 236,057 most recent pre-cutoff conclusions. We repeated the comparison using an impact-based set selected by Relative Citation Ratio (RCR), comprising 236,057 conclusions across 2011--2023. ForeSCITE generated hypotheses by decoding temporally extrapolated embeddings at different values of $\alpha$. For downstream hypothesis generation, we selected the operating regime $\alpha<1.0$ using paired semantic similarity, effective-rank loss and shared UMAP geometry (Methods, Section~\ref{sec:alpha_selection}; Appendix Fig.~\ref{fig:cav_alpha_analysis}).

\begin{figure}[!htbp]
    \centering
    \small
    \begin{minipage}[t]{0.48\textwidth}
        \textbf{a}\par\vspace{0.2ex}
        \includegraphics[width=\linewidth]{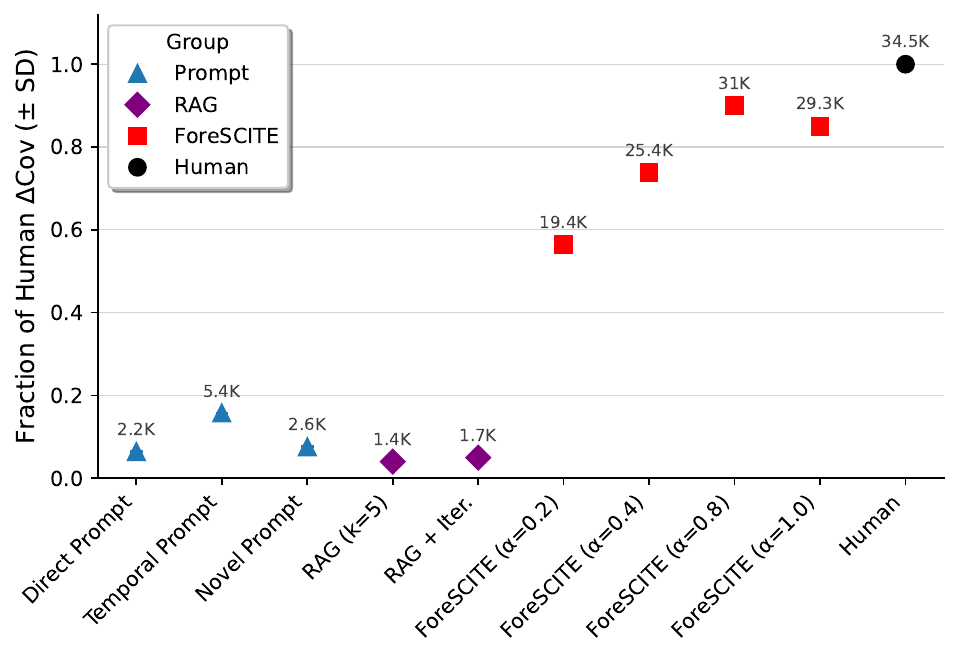}
    \end{minipage}
    \hfill
    \begin{minipage}[t]{0.48\textwidth}
        \textbf{b}\par\vspace{0.2ex}
        \includegraphics[width=\linewidth]{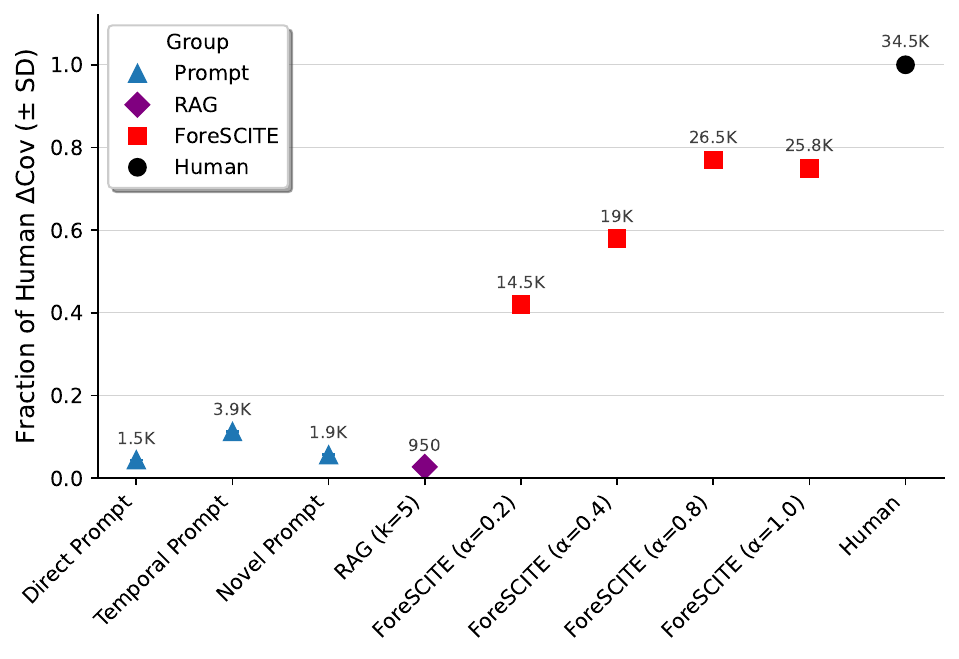}
    \end{minipage}\\[1ex]
    \begin{minipage}[t]{0.48\textwidth}
        \textbf{c}\par\vspace{0.2ex}
        \includegraphics[width=\linewidth]{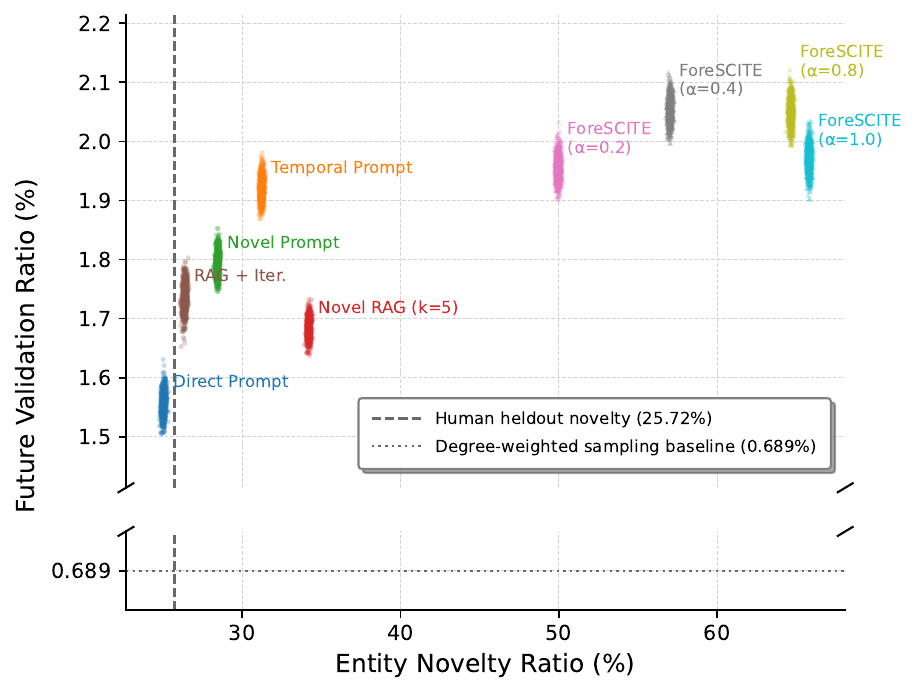}
    \end{minipage}
    \hfill
    \begin{minipage}[t]{0.48\textwidth}
        \textbf{d}\par\vspace{0.2ex}
        \includegraphics[width=\linewidth]{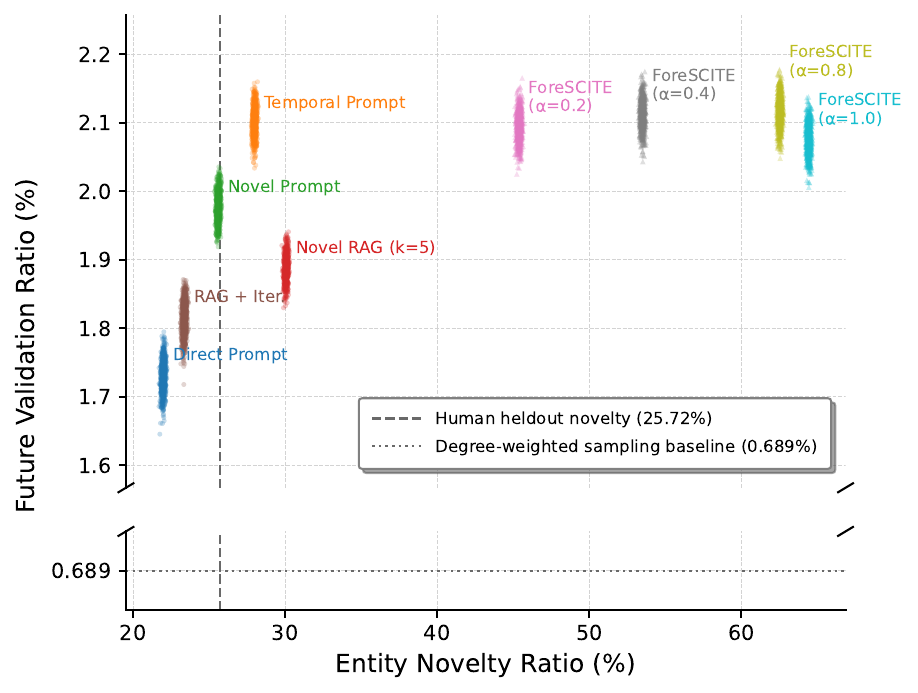}
    \end{minipage}
    \caption{\textbf{ForeSCITE expands semantic coverage and generates novel entity pairs that align with future literature under two reference-selection strategies.} \textbf{a,b,} Semantic space growth, $\Delta \text{Cov}$, measured by the \#Circles metric at radius $r=0.7$ for recency-based and impact-based reference conclusions, respectively. Coverage is reported as a fraction of the coverage achieved by the 236,057 human-authored post-cutoff conclusions; error bars denote s.d. Labels above the points report the corresponding absolute coverage values. \textbf{c,d,} Bootstrap distributions of the Entity Novelty Ratio and Future Validation Ratio calculated from co-occurring biomedical entity pairs for the recency-based and impact-based reference conclusions, respectively. For each method and selection strategy, hypotheses were resampled with replacement 1,000 times and both ratios were recomputed for every sample (9,000 bootstrap samples per panel). Each point represents one bootstrap resample. The vertical dashed line marks the human-written post-cutoff Entity Novelty Ratio of 25.72\%, calculated from 161,625 novel pairs among 628,434 pairs extracted from 236,057 conclusions. The horizontal dotted line marks the degree-weighted sampling baseline for the Future Validation Ratio (0.689\%); the vertical axis is broken to display this baseline alongside the generation results. Both the recency-based and impact-based sets contain 236,057 pre-cutoff conclusions (Appendix Table~\ref{tab:reference_selection}). Across both analyses, ForeSCITE generations show a tradeoff between increasing entity novelty and maintaining alignment with future literature.}
    \label{fig:coverage_symbolic_eval}
\end{figure}

Using the density-based \#Circles metric~\cite{xieHowMuchSpace2023}, we measured semantic space growth, $\Delta \text{Cov}$, at radius $r=0.7$ and normalized each result by the coverage attained by the 236,057 human-authored post-cutoff conclusions. In both selection analyses, prompt-based and retrieval-augmented baselines covered less of the future semantic space than ForeSCITE (Fig.~\ref{fig:coverage_symbolic_eval}a,b). At $\alpha=0.8$, ForeSCITE generations reached approximately 90\% of human coverage with recency-based references and approximately 77\% with impact-based references.

To evaluate whether this geometric expansion corresponded to biologically meaningful relational novelty, we used the offline PubTator 3.0 tagging pipeline~\cite{weiPubTator30AIpowered2024} to identify biomedical entities in generated hypotheses, constructed pairs of entities co-occurring within each hypothesis, and compared these pairs with pre- and post-cutoff reference sets. For each reference-selection strategy, we resampled the hypotheses from each method with replacement 1,000 times and computed the Entity Novelty Ratio and Future Validation Ratio for every bootstrap sample (Fig.~\ref{fig:coverage_symbolic_eval}c,d).

To contextualize the Entity Novelty Ratio, we evaluated the 236,057 human-written post-cutoff conclusions using the same novelty definition. Of the 628,434 extracted biomedical entity pairs, 161,625 were absent from the pre-cutoff training corpus, yielding a human-written Entity Novelty Ratio of 25.72\%. This corpus-level value provides an empirical reference for the degree of relational novelty in generated hypotheses and is shown in both reference-selection analyses (Fig.~\ref{fig:coverage_symbolic_eval}c,d).

To contextualize the Future Validation Ratio, we also included two chance baselines that sampled unseen pairs from the pre-cutoff vocabulary of 317,547 entities and scored them against the same novel post-cutoff co-occurrence set used for the generation methods. Uniform sampling yielded an analytic expected rate of 0.00095\% (99\% null sampling interval, 0--0.003\%), whereas degree-weighted sampling, with endpoint probabilities proportional to pre-cutoff training degree, yielded 0.689\% (99\% null sampling interval, 0.641--0.727\%). These intervals summarize variability across 1,000 random sets of 200,000 unique unseen pairs. The degree-weighted baseline provides a reference that accounts for the greater connectivity of some entities in the historical literature, rather than treating every unseen combination as equally likely.

Across both reference-selection strategies, ForeSCITE occupied regions with higher novelty and future validation than the prompt-based and RAG baselines. Increasing $\alpha$ shifted ForeSCITE generations toward greater entity novelty, although the highest validation ratios occurred at intermediate extrapolation strengths, illustrating a tradeoff between maximizing novelty and retaining alignment with future literature. These results indicate that extrapolation along research trend directions expands the hypothesis space toward novel biomedical entity pairs subsequently observed in the literature, with $\alpha$ providing explicit control over that balance.

\subsection{ForeSCITE identifies novel and plausible hypotheses on CAR-T therapy}

\begin{figure}[!htbp]
    \centering
    \setlength{\parskip}{0pt}
    \begin{minipage}[t]{\textwidth}
        \textbf{a}\par\vspace{0.5ex}
        \includegraphics[width=\linewidth]{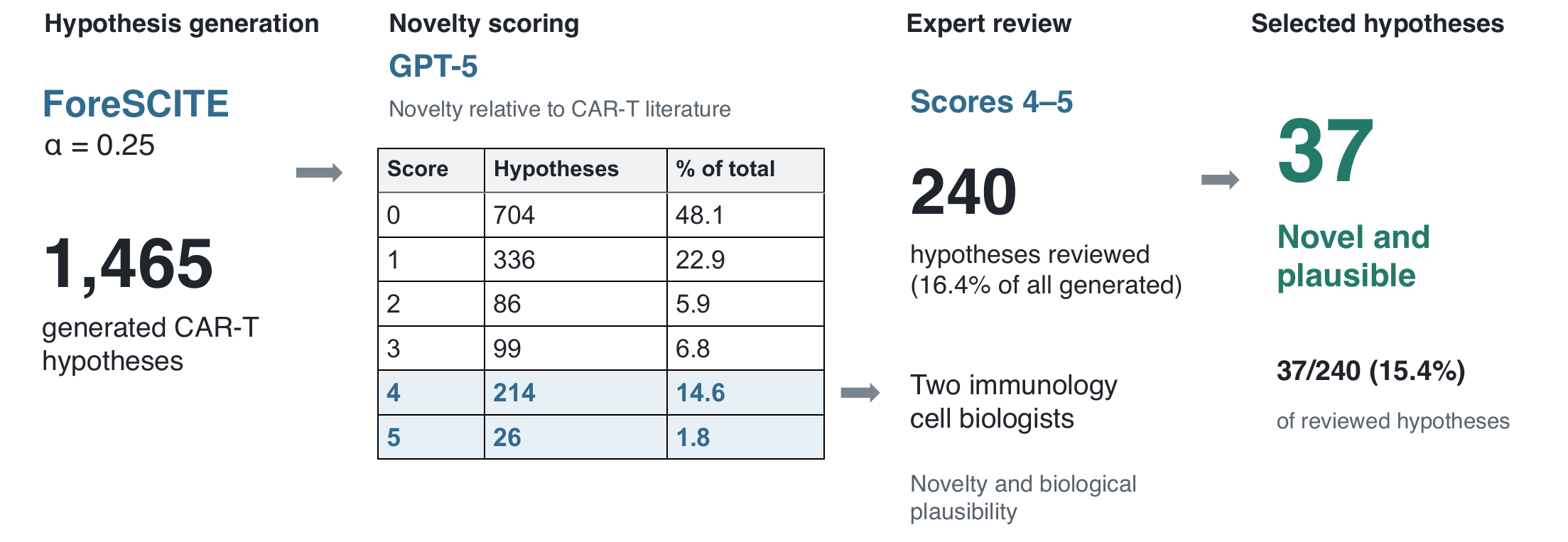}
    \end{minipage}
    \par\vspace{1.5ex}
    \begin{minipage}[t]{\textwidth}
        \textbf{b}\par\vspace{0.5ex}
        \includegraphics[width=\linewidth]{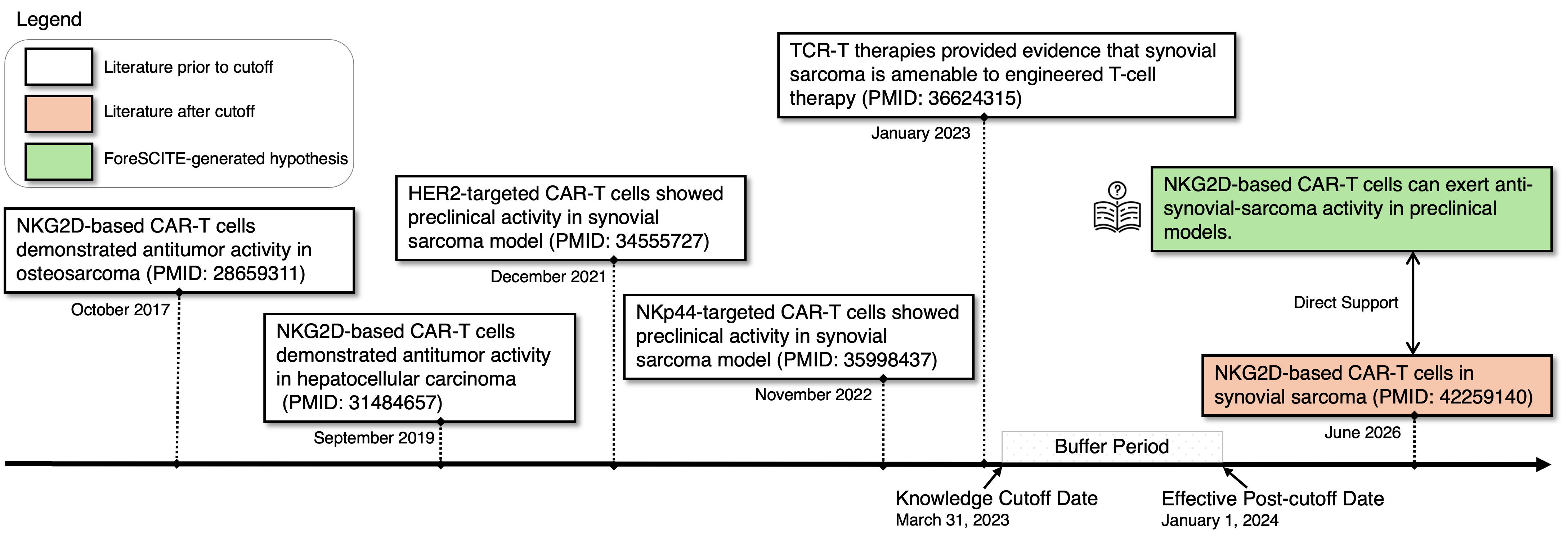}
    \end{minipage}
    \caption{\textbf{Expert assessment and temporal evidence for ForeSCITE-generated CAR-T hypotheses.} \textbf{a,} ForeSCITE generated 1,465 CAR-T hypotheses with $\alpha=0.25$. GPT-5 assessed novelty on an ordinal scale from 0 (very low) to 5 (very high); counts and percentages are shown for each score, with scores 4 and 5 highlighted. Two immunology cell biologists reviewed all 240 hypotheses with scores of 4 or 5 (16.4\% of all generated hypotheses) for novelty and biological plausibility. Expert assessment identified 37 hypotheses as both novel and plausible: 15.4\% of the reviewed subset and 2.5\% of all generated hypotheses. \textbf{b,} Temporal evidence for the NKG2D-based CAR-T hypothesis in synovial sarcoma. Pre-cutoff studies supported NKG2D-directed activity in other tumors and engineered T-cell approaches in synovial sarcoma. White, green and peach boxes denote pre-cutoff literature, the ForeSCITE-generated hypothesis and post-cutoff literature, respectively. The later preclinical study provides direct support for the disease-specific hypothesis. The timeline marks the March 31, 2023 knowledge cutoff and a buffer period ending January 1, 2024.}
    \label{fig:car_t_expert_review}
\end{figure}

Having shown that temporal extrapolation expands the semantic and relational coverage of generated hypotheses, we next assessed whether it produces candidates judged by domain experts to be both novel and biologically plausible. We used CAR-T therapy as an example topic because of its impact in immunology and the clinical application of cancer immunotherapy. ForeSCITE generated 1,465 CAR-T hypotheses at $\alpha=0.25$, which we prioritized using GPT-5 novelty scores before assessment by two immunology cell biologists (Methods, Section~\ref{sec:car_t_expert_assessment}; Fig.~\ref{fig:car_t_expert_review}a).

GPT-5 assigned novelty scores of 0, 1, 2, 3, 4 and 5 to 704 (48.1\%), 336 (22.9\%), 86 (5.9\%), 99 (6.8\%), 214 (14.6\%) and 26 (1.8\%) hypotheses, respectively. All 240 hypotheses with scores of 4 or 5 (16.4\% of the generated set) proceeded to expert review. The two immunology cell biologists identified 37 hypotheses as both novel and biologically plausible, corresponding to 15.4\% of the reviewed subset and 2.5\% of all generated hypotheses (Fig.~\ref{fig:car_t_expert_review}a). The remaining 203 reviewed hypotheses were not retained; aggregate review outcomes did not distinguish failures of novelty from failures of plausibility. Hypotheses with scores of 0--3 were not included in this screening review, so the 2.5\% figure represents the proportion of all generated hypotheses selected by this workflow, not an estimate of the overall prevalence of novel and plausible hypotheses.

We further examined five selected hypotheses spanning receptor targeting, cell-surface engineering, transcriptional regulation and metabolic support. The expert justifications below describe their biological rationale, novelty relative to earlier literature and subsequent published evidence where available.

\subsubsection*{Hypothesis 1 (‘NKG2D-based CAR-T cells’ + ‘synovial sarcoma’)}

This hypothesis connects ‘NKG2D-based CAR-T cells’ with ‘synovial sarcoma’, representing a novel extrapolation of multi-ligand targeting to a refractory mesenchymal tumor. Although synovial sarcoma demonstrates clinical responsiveness to HLA-restricted TCR-T regimens~\cite{hongAutologousCellTherapy2023}, CAR-T therapies circumvent HLA dependency to broaden clinical applicability. The activating receptor NKG2D binds multiple stress-induced ligands, affording wider target coverage than single-antigen designs. Evidence prior to March 2023 supported alternative CAR-T targets (HER2, NKp44) in synovial sarcoma~\cite{murayamaEffectiveness41BBcostimulatedHER2targeted2021,murayamaNKp44basedChimericAntigen2022}, alongside NKG2D-directed efficacy in osteosarcoma and hepatocellular carcinoma~\cite{fernandezMemoryCellsExpressing2017,sunEradicationHepatocellularCarcinoma2019}. Nevertheless, applying an NKG2D strategy specifically to synovial sarcoma remained unaddressed in published literature. ForeSCITE bridged this knowledge gap by hypothesizing disease-specific preclinical activity for NKG2D CAR-T cells. In June 2026, Miyazaki et al. validated this prediction~\cite{miyazakiNKG2DbasedCARTCells2026}, demonstrating NKG2D-ligand expression in synovial sarcoma models, robust cytokine release and CD107a degranulation, and marked in vivo tumor control in mouse xenografts. Figure~\ref{fig:car_t_expert_review}b summarizes the temporal evidence, placing the relevant pre-cutoff findings alongside the generated hypothesis and subsequent preclinical evidence. These results substantiate ForeSCITE’s utility in synthesizing disparate immunological findings into actionable, disease-specific therapeutic strategies.

\subsubsection*{Hypothesis 2 (‘PEGylated CAR-T cells’)}

Severe systemic toxicities, principally cytokine release syndrome (CRS) and neurotoxicity, frequently constrain the therapeutic window of CD19 CAR-T cells in B-cell malignancies. Addressing this bottleneck, the ForeSCITE-generated concept of “PEGylated CAR-T cells” introduces a materials-engineering strategy to decouple efficacy from toxic hyperinflammation. Direct surface PEGylation of living CAR-T cells to mitigate systemic adverse events had not been reported prior to March 2023, marking an unmapped therapeutic territory. ForeSCITE predicted that a PEGylated platform could yield cell products with superior safety without compromising therapeutic intent. A landmark September 2023 study corroborated this directional insight~\cite{gongSituPEGylationCAR2023}, showing that in situ PEGylation installs a temporary polymeric spacer on T cells to buffer hyperactive interactions with tumor cells and monocytes, thereby preventing acute monocyte-driven CRS. As CAR-T cells expand in vivo, the PEG coating naturally dilutes, permitting a delayed, full recovery of antitumor cytotoxicity past the critical window for acute toxicity. This empirical confirmation underlines ForeSCITE's capacity to leverage cross-disciplinary bio-surface engineering for safer cell therapies.

\subsubsection*{Hypothesis 3 (‘PRDM1/TCF3 double knockdown’)}

Rapid loss of stemness and epigenetic commitment to terminal exhaustion impede the durability of CD8+ adoptive T-cell therapies, including CAR-T therapies. To overcome these redundant dysfunction programs, ForeSCITE proposed ‘simultaneous knockdown of PRDM1 and TCF3’ as a multi-target transcriptional intervention. PRDM1 (encoding BLIMP1) drives terminal T-cell differentiation; although PRDM1 ablation preserves TCF1-dependent stemness, its solo disruption triggers compensatory, NFAT-driven transcriptional feedback (upregulating NR4A3) that impairs solid tumor efficacy, necessitating multi-gene knockout strategies~\cite{jungBLIMP1NR4A3Transcription2022}. Before March 2023, despite foundational knowledge on TCF3 (E2A) in T-cell biology, targeting TCF3 to counteract CAR-T exhaustion remained unexplored. Subsequently, a March 2024 study revealed that TCF3 complexes with the histone demethylase LSD1 at E-box elements to epigenetically preserve Slamf6+ progenitor-exhausted T cells ($\text{Tex}^\text{prog}$) and drive TCF1 expression~\cite{liId2EpigeneticallyControls2024}. ForeSCITE predicted that dual knockdown of PRDM1 and TCF3 network components could synergistically alleviate BLIMP1-driven exhaustion while stabilizing TCF1/Slamf6 epigenetic states, preventing single-gene compensatory resistance. Although the in vivo persistence and cytotoxic kinetics of PRDM1/TCF3 double-knockdown CAR-T cells await direct experimental validation, these insights offer a rational multiplex editing roadmap for next-generation immunotherapies.

\subsubsection*{Hypothesis 4 (‘CD40L/FOXP3 co-expression’ + ‘advanced renal cell carcinoma’)}

Severe metabolic restriction, physical immune exclusion, and antigen loss collectively enforce T-cell dysfunction in recalcitrant solid tumor microenvironments. The ForeSCITE-suggested combination of ‘CD40L/FOXP3 co-expression’ for ‘advanced renal cell carcinoma’ introduces a dual-action genetic blueprint to counter these multifaceted resistance mechanisms. Constitutive CD40L expression enhances CAR-T proliferation and TH1 cytokine release while upregulating costimulatory ligands to boost overall tumor niche immunogenicity~\cite{curranEnhancingAntitumorEfficacy2015}, simultaneously licensing antigen-presenting cells to recruit endogenous T cells and counteract antigen-loss escape~\cite{kuhnCD40LigandModifiedChimeric2019}. Concurrently, while FOXP3 is conventionally assigned to regulatory lineage specification, recent findings show that it drives a distinct metabolic reprogramming in effector CAR-T cells; by upregulating lipid oxidation, FOXP3 confers vital metabolic adaptability in hypoxic, nutrient-scarce microenvironments without imparting suppressive function~\cite{niuFoxp3ConfersLongterm2025}. Pairing these factors creates a synergistic architecture: CD40L acts as a microenvironment disrupter and immune recruiter, whereas FOXP3 metabolically armors CAR-T cells against exhaustion. While the in vivo synergy of this dual-engineered configuration in renal cell carcinoma remains to be empirically verified through targeted preclinical studies, it underscores the ForeSCITE’s capacity to formulate innovative, multi-pathway genetic cell modifications.

\subsubsection*{Hypothesis 5 (‘N-acetylcysteine supplementation’ + ‘CAR-T cell persistence in lymphoma’)}

In aggressive B-cell lymphomas, reactive oxygen species (ROS) accumulation and activation-induced cell death (AICD) frequently impair the expansion and longevity of CD19 CAR-T cells. The ForeSCITE-proposed integration of N-acetylcysteine (NAC) co-administration offers an accessible redox-metabolic strategy to preserve cellular fitness. Prior to March 2023, preclinical studies established that NAC mitigates TCR-driven DNA damage signaling (ATM-dependent p53 phosphorylation and \ensuremath{\gamma}H2AX upregulation) to prevent AICD~\cite{scheffelEfficacyAdoptiveTcell2016}. Additionally, dampening ROS exposure during naive T-cell priming stalls terminal differentiation, favoring the enrichment of self-renewing stem cell memory T (Tscm) cells with superior in vivo persistence and antitumor efficacy upon CAR targeting~\cite{pilipowAntioxidantMetabolismRegulates2018}. Nevertheless, translating NAC as an in vivo adjunct during commercial CAR-T therapy in patients represented an untested clinical paradigm. A prospective 2025 Phase I trial in relapsed/refractory lymphoma patients receiving axicabtagene ciloleucel CAR T cells~\cite{shahInfusionalNAcetylcysteineEnhances2025} subsequently verified this prediction, demonstrating that concurrent NAC infusion safely elevated 1-week CAR-T persistence (29.7\% vs. 4.6\% in controls, p=0.026) and reduced PD-1 expression without additive toxicity. These clinical outcomes demonstrate the ForeSCITE’s efficacy in prospectively translating preclinical metabolic concepts into viable clinical protocols.

\section{Discussion}

This study introduces a different computational formulation of biomedical hypothesis generation: instead of asking a language model to propose an idea from retrieved evidence, ForeSCITE estimates how scientific claims have moved through a semantic space and decodes extrapolated coordinates into natural language. Three findings support this formulation. First, conclusion embeddings within thematic areas contained reproducible temporal structure, allowing a dominant early-to-late direction to be estimated across 407 clusters. Second, moving embeddings along these directions expanded both semantic coverage and biomedical entity-pair novelty substantially more than direct prompting or retrieval-augmented generation, while the temporal step size provided explicit control over the balance between displacement and diversity. Third, an expert-review workflow identified 37 CAR-T hypotheses as both novel and biologically plausible, while five selected examples illustrated biological rationales, potential therapeutic applications and differing levels of subsequent literature support. Together, these results suggest that temporal extrapolation can serve as a scalable hypothesis-proposal mechanism, converting the historical motion of the literature into candidates for subsequent scientific evaluation.

ForeSCITE occupies a distinct position between temporal literature-based discovery and recent LLM co-scientists. Temporal LBD methods have shown that evolving graphs contain predictive information, but generally express a hypothesis as a future link between predefined concepts~\cite{zhouLearningTemporalDifference2022,zhouGeneratingBiomedicalHypothesis2024,shariatmadariConceptDriftLeveragingSpatial2025}. ForeSCITE instead models complete scientific conclusions, preserving contextual attributes such as directionality, experimental setting and proposed consequence that are difficult to represent as a single edge. Conversely, agentic systems such as Co-Scientist, Robin and the Virtual Lab use retrieval, deliberation, critique or experimental feedback to search a problem defined by a scientist~\cite{gottweisAcceleratingScientificDiscovery2026,ghareebMultiagentSystemAutomating2026,swansonVirtualLabAI2025}. ForeSCITE does not replace these capabilities. Its contribution is to supply a data-derived direction before deliberation begins, where the literature is treated not only as evidence to retrieve, but as a dynamical record from which a search direction can be estimated. Accordingly, our baseline comparisons establish an advantage over matched prompt-only and retrieval-augmented generation conditions for large-scale trajectory extrapolation; they do not establish superiority over full multi-agent systems optimized for a specific research objective.

The temporal step size clarifies what this form of generation is doing. Small steps produced hypotheses close to known conclusions, whereas larger steps increased semantic and relational novelty but eventually compressed the generated distribution into a lower-rank region. The strongest future alignment occurred at intermediate rather than maximal displacement. This pattern is consistent with an ``adjacent possible''~\cite{kauffmanInvestigations2000} interpretation of scientific forecasting. A useful prediction must depart from what is already known, but remain connected to a trajectory for which the historical literature provides sufficient support. In this view, $\alpha$ is not merely a decoding hyperparameter; it specifies a search policy along a continuum from conservative continuation to speculative extrapolation. The onset of space collapse at larger values further indicates that extrapolation cannot be increased without cost and motivates selecting $\alpha$ from several criteria, including semantic displacement and population diversity, rather than novelty alone.

The human-written post-cutoff Entity Novelty Ratio of 25.72\% provides an empirical scale for relational novelty; a higher novelty ratio alone does not establish greater scientific utility. The absolute Future Validation Ratio requires similarly careful interpretation. Even for the best-performing ForeSCITE settings, only about 2\% of generated entity pairs were novel pairs observed in the post-cutoff corpus. This percentage should not be read as showing that the remaining hypotheses were false. The denominator contains open-ended proposals spanning many entities and disease contexts, whereas the holdout window is short relative to the time required to prioritize, fund, conduct and publish biomedical research. Moreover, only a small and unknown fraction of scientifically plausible hypotheses will be investigated during any fixed interval. The metric is therefore better interpreted as a conservative future re-observation rate under incomplete ascertainment, and its primary evidence is comparative. ForeSCITE increased this rate while also generating substantially more novel pairs than the matched prompt-based and retrieval-augmented baselines. The two chance baselines provide additional context, where uniform sampling of unseen pairs yielded an expected rate of 0.00095\%, whereas degree-weighted sampling yielded 0.689\%. The approximately 2\% rate at the best-performing ForeSCITE settings exceeds both expectations, suggesting that future overlap is not explained only by arbitrary pairing or preferential sampling of historically well-connected entities under these null models. These references are nevertheless restricted to unseen pairs within the pre-cutoff vocabulary and do not control for all differences in topic, entity type or research activity. It remains necessary to establish practical predictive value.

The five selected CAR-T hypotheses complement the population-level metrics by illustrating how generated ideas can connect existing findings across disease settings, engineering strategies and biological mechanisms. The expert justifications describe subsequent support for NKG2D-based CAR-T activity in synovial sarcoma, transient PEGylation to mitigate toxicity and N-acetylcysteine supplementation to support persistence. They also identify the PRDM1/TCF3 and CD40L/FOXP3 combinations as proposals that still require direct experimental evaluation. These differences emphasize the need to distinguish support for individual biological components from validation of a complete generated hypothesis. Because the cases are selected illustrations assessed with access to later literature, they do not establish a prospective success rate.

Several limitations follow from the current representation of scientific change. A single learned research trend direction summarizes the historical change within each cluster. Although this first-order approximation is stable and interpretable, research fields may branch, converge, reverse after negative findings or change abruptly after a new technology. The cluster-level model may also obscure local trajectories followed by smaller subfields and may be less reliable in sparse or rapidly reorganizing domains. Local vector fields, time-varying or nonlinear trajectory models, and mixtures of multiple future directions could represent these dynamics more faithfully. Such models should also quantify uncertainty. Clusters with weak temporal separation or projected points far from the observed manifold should yield lower-confidence hypotheses or abstain from generation.

ForeSCITE also inherits biases and information loss from its data and evaluation pipeline. PubMed conclusions preferentially represent published, predominantly positive findings and do not capture unsuccessful experiments, unpublished work or all relevant clinical and molecular evidence. Conclusion extraction, embedding geometry, clustering, ELM decoding and automated entity annotation can each introduce errors, and entity co-occurrence does not by itself establish a mechanistic relation. The semantic coverage metric measures expansion of a particular embedding space rather than scientific utility, while later publication can reflect popularity, funding and data availability as well as biological truth. In addition, the present LLM baselines share one base model and a limited set of prompts and retrieval strategies; broader comparisons should control model scale and inference budget and should include contemporary agentic systems where their input assumptions are compatible. Finally, evaluation on post-cutoff holdout literature tests temporal generalization, but temporal separation alone cannot eliminate every source of model-pretraining or selection leakage. A prospective evaluation, with corpora, models, generated hypotheses and ranking rules frozen before new evaluation evidence becomes available, is needed to measure calibration and real-world yield.

These limitations point to a complementary role for trajectory-based and deliberative systems. ForeSCITE is likely to be most effective at systematically enumerating trajectory-consistent next steps, whereas agentic search may be better suited to deliberately constructing cross-field combinations, challenging dominant assumptions and developing experimental plans. A hybrid system could use ForeSCITE to propose diverse, uncertainty-aware future coordinates and then apply retrieval, multi-agent critique, causal analysis and feasibility constraints to rank and refine the decoded hypotheses. The most important next test is prospective: timestamp and preregister generated candidates, evaluate them blindly with domain experts, release calibrated rankings, and follow both literature emergence and experimental outcomes. Ultimately, scientific hypotheses range from incremental extensions to paradigm-changing discontinuities. ForeSCITE addresses a tractable and consequential part of that spectrum by learning the direction of the adjacent possible; integrating this signal with mechanisms for exploratory recombination may provide a broader foundation for autonomous scientific discovery.

\section{Methods}

As summarized in Fig.~\ref{fig:framework}, ForeSCITE is organized as a framework for converting historical biomedical literature into temporally extrapolated hypotheses. First, we prepare a relation-enriched corpus of PubMed conclusions and embed each conclusion into a shared latent space. Second, we train an Embedding Language Model (ELM) to decode points in this space back into conclusion-like biomedical text. Third, we estimate cluster-specific research trend directions that summarize changes from earlier to more recent published conclusions within thematic research areas. Finally, we project known pre-cutoff conclusion embeddings along these temporal directions and decode the resulting future coordinates into natural language hypotheses. These components jointly enable ForeSCITE to move from grounded literature claims to plausible hypotheses that extend beyond the observed training corpus.

\subsection{Preparing literature data \& embedding space} 
We constructed a comprehensive biomedical literature corpus using approximately 39 million records from the PubMed File Transfer Protocol server. This dataset includes the 2025 annual baseline and all daily update files through August 05, 2025. To ensure our dataset reflects experimental outcomes and scientific hypotheses, we restricted our selection to records containing abstracts. For each remaining record, we extracted the conclusion section by parsing structured XML tags or, in the case of unstructured abstracts, employing a sentence classification model~\cite{huMoreGeneralizableAccurate2023} to identify conclusion-specific text. To ground these conclusions in verifiable biological knowledge, we aligned the extracted conclusion texts with PubTator 3.0~\cite{weiPubTator30AIpowered2024} annotations. We retained only those conclusions containing at least one PubTator-defined biological relation, such as gene-gene, chemical-gene and gene-disease associations. This filtering process yielded a final dataset of 3,305,718 conclusions associated with 7,749,555 PubTator relations. We adopted a temporal split framework and partitioned the data based on the knowledge cut-off date, March 31, 2023, of our base Large Language Model (LLM), Meta-Llama-3-8B-Instruct. Data published prior to this date was used for training (3,005,395 conclusions) and validation (64,266 conclusions). Data published after this date was reserved as a holdout set (236,057 conclusions) to evaluate and benchmark the model's ability to extrapolate toward future scientific findings.

For ForeSCITE extrapolation, we evaluated two strategies for selecting pre-cutoff reference conclusions. The recency-based set comprised the 236,057 most recent eligible conclusions before the cutoff date. The impact-based set comprised eligible conclusions published from 2011 through 2023 with a Relative Citation Ratio (RCR) greater than 2, using RCR values from the iCite database. RCR is a field- and time-normalized citation metric; an RCR greater than 2 indicates that a paper receives more than twice as many citations per year as the median NIH-funded paper in its field. The year counts for both selection strategies sum to 236,057 (Appendix Table~\ref{tab:reference_selection}). All conclusions were encoded into a 1,024-dimensional latent space using the BAAI/bge-large-en-v1.5~\cite{xiaoCPackPackedResources2024} embedding model.

\subsection{Training embedding language model} 
We trained an ELM~\cite{tennenholtzDemystifyingEmbeddingSpaces2024,ondov-etal-2026-coreelm} to function as an embedding inversion operator, translating high-dimensional embeddings back into human-readable scientific hypotheses. Similar to the visual encoders used in vision language models, the ELM architecture augments an additional encoder to the base LLM, treating the embedding space as a distinct modality. Let $E_{emb}:\mathcal{X}\rightarrow\mathcal{Z}_{emb}$ represents the mapping of a sequence of tokens $\mathcal{X}$ to the 1,024-dimensional embedding space. Our target model, $\mathcal{M}_{tgt}$, is designed to interpret these external embeddings by aligning them with the internal token space of a base LLM,  $\mathcal{M}$. We utilized a two-layer multilayer perceptron as an adapter to project the conclusion embeddings into the token space of the base LLM:

$$W_1(\sigma\left(W_0\mathcal{Z}_{emb}+b_0\right)+b_1,$$

where $W$ and $b$ represent learnable weights and biases, and $\sigma$ denotes a non-linear activation function. During the training phase, we kept the token embedding layer of the base LLM frozen and only optimized the adapter and the transformer layers. To maintain computational efficiency and prevent catastrophic forgetting, we employed Low-Rank Adaptation (LoRA)~\cite{huLoRALowRankAdaptation2021} to optimize query and key weights in transformer layers.

The ELM was aligned to the embedding space of biomedical conclusions by reconstructing the original natural language conclusion from its embedding representation. This ensures the model captures the full semantic nuance of the literature. By training on this task, the ELM learns to map the geometric proximity of vectors to the logical proximity of biomedical conclusions and concepts. This alignment allows the framework to ingest a hypothetical embedding and decode it into a hypothesis that has not yet appeared in the training corpus.

\subsection{Estimating research trend directions}
To capture the temporal trajectories of distinct biomedical domains, we identified temporal development trends at the thematic level. Given the high dimensionality and sparsity of the 1,024-dimensional conclusion embedding space, we first reduced the 3,005,395 pre-cutoff training embeddings to 10 components using Uniform Manifold Approximation and Projection (UMAP)~\cite{mcinnesUMAPUniformManifold2018} to facilitate stable density-based clustering. We then applied Hierarchical Density-Based Spatial Clustering of Applications with Noise~\cite{mcinnesHdbscanHierarchicalDensity2017} using Euclidean distance and a minimum cluster size of 500. We selected this value because it achieved a strong density-based clustering validation (DBCV) score while assigning fewer points to noise than a minimum cluster size of 250 (Appendix Fig.~\ref{fig:hdbscan_min_cluster_size}). This procedure identified 407 thematic clusters $K$ representing specialized areas of biomedical research.

For each cluster $k\in K$, we estimated a research trend direction in the original 1,024-dimensional embedding space, following the concept activation vector approach~\cite{kimInterpretabilityFeatureAttribution2018}. Each direction is a unit vector pointing from earlier toward more recent conclusion representations. We calculated the 25th and 75th percentiles of publication year within the cluster, denoted $q_{1,k}$ and $q_{3,k}$. Conclusions published in years less than or equal to $q_{1,k}$ were labeled early, and those published in years greater than or equal to $q_{3,k}$ were labeled late; conclusions between the two thresholds were excluded from classifier training.

We deliberately used a linear support-vector classifier (\texttt{LinearSVC})~\cite{fanLIBLINEARLibraryLarge2008} as a first-order model of temporal change. To motivate this choice, consider an unknown smooth temporal decision score $f_k(x)$ in the embedding space. Around a reference point $x_0$, its first-order Taylor approximation is

$$f_k(x)\approx f_k(x_0)+\nabla f_k(x_0)^{T}(x-x_0).$$

The learned linear SVC score can be written in the same form,

$$d_k(x)=w_k^{T}x+b_k=d_k(x_0)+w_k^{T}(x-x_0).$$

Accordingly, $w_k=\nabla d_k(x)$ is the constant gradient of the learned decision score and the normal vector of the maximum-margin hyperplane separating early from late conclusions. It therefore provides a first-order estimate of the direction in which temporal class evidence increases most rapidly within cluster $k$. This interpretation does not assume that the true evolution of the literature is linear; rather, the linear model intentionally compresses its dominant early-to-late trend into a single stable and interpretable direction that can be used for vector extrapolation.

We evaluated temporal separability using five-fold cross-validation. The per-cluster early--late classification accuracy reported in Fig.~\ref{fig:temporal_structure}b,c is the mean accuracy across the five folds. For this analysis, cluster size is defined as the combined number of early and late conclusions used for classifier training. To summarize the association between data availability and temporal separability, we regressed cross-validation accuracy on the log-transformed cluster size and calculated a 95\% confidence band. Finally, we refitted the classifier using all early and late examples in each cluster~$k$ and normalized its coefficient vector $w_k$ to unit length,

$$\upsilon_k=\frac{w_k}{\lVert w_k\rVert_2}.$$

The sign was oriented so that $\upsilon_k$ points toward the late class, representing the direction from earlier to more contemporary conclusions.

\subsection{Generating hypotheses by temporal extrapolation}

We used the learned research trend directions to guide hypothesis generation in the embedding space. For each reference conclusion, we performed temporal extrapolation along the direction learned for its thematic cluster. The reference embedding $z \in \mathcal{Z}_{emb}$ represents a known biomedical conclusion selected from the pre-cutoff training set. The extrapolated embedding, $\hat{z}$, is calculated as:

$$\hat{z}=z+\alpha\upsilon_k,$$

where the hyperparameter $\alpha$ controls displacement in embedding space rather than elapsed calendar time. This process shifts the semantic representation of a known conclusion along the research trend direction to propose a possible continuation of the historical pattern. To translate these high-dimensional projections into human-readable text, we employed the trained ELM, $\mathcal{M}_{tgt}$, as a decoding interface. The model ingests $\hat{z}$ and decodes it into a natural-language hypothesis, $\hat{x}=\mathcal{M}_{tgt}(\hat{z})$, with the syntactic structure of a biomedical conclusion.

\subsection{Publication-year regression and parametric temporal visualization}
\label{sec:temporal_visualization_methods}

We used the thematic partition of the 3,005,395 pre-cutoff conclusion embeddings into 407 clusters described above to examine temporal information within clusters. Figure~\ref{fig:temporal_structure}a shows three illustrative clusters: Molecular Hematologic Malignancy Research, Diabetic Vascular and Renal Complications, and Rare Monogenic Disease Genetics. The analyses use the original 1,024-dimensional embeddings. Publication time is represented numerically as publication year, and pre- and post-cutoff groups follow the March 31, 2023 corpus split.

\subsubsection{Linear prediction of publication year}

For each displayed cluster $k$, we fitted a linear regression from a conclusion embedding $z\in\mathbb{R}^{1024}$ to its publication year $y$,
\[
    \widehat{y}_{\mathrm{lin},k}(z)=\beta_k^{T}z+b_{\mathrm{lin},k},
\]
where $\beta_k$ and $b_{\mathrm{lin},k}$ are the fitted regression coefficients and intercept. The top-row visualizations group predicted years by actual publication-period bins and show their distributions. Pre- and post-cutoff conclusions are distinguished by color to compare temporal ordering across the corpus boundary.

\subsubsection{Supervised two-dimensional mapping and decoded movement}

For the movement visualizations, we fitted a cluster-specific MLP to predict publication year from the same 1,024-dimensional conclusion representation. The network architecture was $1024\rightarrow16\rightarrow8\rightarrow2\rightarrow1$, with a Leaky ReLU activation at the two-dimensional bottleneck. Denoting the output of this penultimate layer by $h_k(z)\in\mathbb{R}^{2}$, the scalar prediction is
\[
    \widehat{y}_{\mathrm{MLP},k}(z)=a_k^{T}h_k(z)+b_{\mathrm{MLP},k}.
\]
We used $h_k(z)$ as the plotting coordinates and colored the background representation by publication year. These coordinates form a supervised, parametric visualization. Once fitted, the same mapping can place newly encoded conclusions into the existing space.

For each displayed cluster, ten sampled recent pre-cutoff conclusions served as references. We applied the extrapolation along the cluster-specific research trend direction described above with $\alpha=0.25$, decoded each projected embedding with the trained ELM $\mathcal{M}_{tgt}$, and encoded the resulting text with the same frozen BAAI/bge-large-en-v1.5 model used for the literature corpus. Writing this encoder as $E_{emb}$, the displayed reference and generated coordinates are
\[
    u_{\mathrm{ref}}=h_k(z),\qquad
    u_{\mathrm{decoded}}=h_k\!\left(E_{emb}\!\left(\mathcal{M}_{tgt}(z+0.25\upsilon_k)\right)\right).
\]

In the bottom row of Fig.~\ref{fig:temporal_structure}a, light-blue markers indicate $u_{\mathrm{ref}}$, red markers indicate $u_{\mathrm{decoded}}$, and a gray line connects each matched pair. The MLP is held fixed when mapping the decoded texts. Thus, the displayed displacement tests whether the decoding--re-embedding cycle retains movement along the learned temporal organization. For the distribution-level comparison in the top row of Fig.~\ref{fig:temporal_structure}a, we also extrapolated the most recent 5\% of pre-cutoff conclusions within each displayed cluster, decoded and re-embedded the generated texts, and passed their embeddings through the fitted linear publication-year predictor. These predictions are displayed as the separate ``Extrapolated'' group.

\subsection{Exploring and selecting the temporal step size}
\label{sec:alpha_selection}

We characterized the effect of $\alpha$ on both the semantic displacement and geometric diversity of generated conclusions. The reference set comprised the 1,024-dimensional embeddings $R\in\mathbb{R}^{n\times1024}$ of the $n=236{,}057$ most recent pre-cutoff conclusions. For each tested $\alpha$, ForeSCITE projected and decoded all reference conclusions, and the generated texts were encoded with the same embedding model to obtain a row-aligned matrix $G_{\alpha}\in\mathbb{R}^{n\times1024}$. After L2-normalizing every row, we calculated the average paired cosine similarity as

$$C_{\alpha}=\frac{100}{n}\sum_{i=1}^{n}\frac{R_i}{\lVert R_i\rVert_2}\cdot\frac{G_{\alpha,i}}{\lVert G_{\alpha,i}\rVert_2}.$$

To provide a domain-level similarity reference, we considered every training cluster containing at least two conclusions. Within each cluster, we L2-normalized the embeddings, randomly permuted their row order, and calculated the mean row-wise cosine similarity between the original and permuted matrices. The horizontal reference in Appendix Fig.~\ref{fig:cav_alpha_analysis}a shows the mean of these cluster-level values, and the shaded band denotes $\pm1$ s.d. across clusters.

We quantified geometric compression using effective rank. For an embedding matrix $M$ with positive singular values $s_1,\ldots,s_m$, we defined $p_i=s_i/\sum_j s_j$ and

$$\operatorname{erank}(M)=\exp\left(-\sum_{i=1}^{m}p_i\log p_i\right).$$

For each $\alpha$, effective-rank retention was $\operatorname{erank}(G_{\alpha})/\operatorname{erank}(R)$, and rank loss was reported as

$$100\left(1-\frac{\operatorname{erank}(G_{\alpha})}{\operatorname{erank}(R)}\right).$$

We additionally visualized distributional changes in a shared two-dimensional coordinate system. We concatenated the 3,005,395 training embeddings with the 236,057 generated-conclusion embeddings from each $\alpha\in\{0.05,0.2,0.4,0.8,1.2\}$ and fitted a single two-component UMAP using the library defaults. Because all historical and generated embeddings were transformed by the same fitted projection, changes in the location and concentration of each generated cloud can be interpreted relative to a common historical landscape.

We selected the operating regime by jointly considering paired cosine similarity, effective-rank loss, and the shared UMAP geometry (Appendix Fig.~\ref{fig:cav_alpha_analysis}). Values below $\alpha=1.0$ kept generated conclusions closer to their references than the typical within-cluster similarity range while limiting rank loss and visible concentration. We therefore anchored downstream hypothesis generation on $\alpha<1.0$; values at or above 1.0 were included as sensitivity analyses to characterize the onset of semantic drift and space collapse.

\subsection{Evaluation \& metric}

\subsubsection{Baselines}
\label{sec:baselines}

We compared ForeSCITE with five LLM baselines under both reference-selection strategies, evaluating all approaches on the same 236,057 conclusions within each strategy (Appendix Table~\ref{tab:reference_selection}). All baselines used the unfine-tuned base model underlying the ELM, Meta-Llama-3-8B-Instruct, without the ELM adapter or LoRA updates. Generations were sampled with temperature 0.6, nucleus-sampling probability $p=0.9$, and a maximum of 256 new tokens. For Direct Prompt, Temporal Prompt, Novel Prompt and Novel RAG, the reference material and generation instruction were provided in a single user turn. RAG + Iter. added a second user turn after the initial model response. In every condition, the model was instructed to return only the predicted conclusion. The complete prompt templates are provided in Appendix~\ref{app:baseline_prompts}.

The first three baselines were prompt-only conditions that received the same reference conclusion and differed only in instructional wording. We included multiple templates because LLM generation can be sensitive to prompt formulation and token length, thereby avoiding reliance on a single prompt as the only prompt-based baseline. Direct Prompt asked for a possible future conclusion that builds on the reference, without explicitly mentioning time, novelty or paraphrasing. Temporal Prompt asked for a hypothetical conclusion that might emerge later from novel hypotheses, emphasizing temporal projection rather than lexical novelty relative to the reference. Novel Prompt explicitly requested a future conclusion based on a novel hypothesis and prohibited repeating or paraphrasing the reference. Together, these conditions represent continuation-, temporal-projection- and novelty-oriented formulations of the same task.

Inspired by SciMON's use of semantic-neighborhood retrieval and iterative refinement to increase novelty~\cite{wangSciMONScientificInspiration2024}, the remaining two baselines incorporated retrieval from the 3,005,395 pre-cutoff training conclusions. For Novel RAG, conclusions were embedded using BAAI/bge-large-en-v1.5 and indexed with a GPU FAISS IVF-PQ index. The index used an L2 coarse quantizer, 4,096 inverted lists, 64 product-quantization subquantizers with 8 bits per subquantizer, and float16 storage. At query time, 64 inverted lists were searched and the five nearest-neighbor conclusions were retrieved. The prompt presented the evaluation conclusion as the primary reference and the retrieved texts as ``Similar Existing Conclusions,'' then requested a novel future conclusion that did not repeat or paraphrase the supplied references.

RAG + Iter. extended Novel RAG with an explicit revision step adapted from SciMON's novelty-improvement instruction to our conclusion-generation task~\cite{wangSciMONScientificInspiration2024}. Its first turn was identical to Novel RAG. The first generated draft was retained in the chat history, and a second user turn asked the model to revise the conclusion so that it differed substantially from both the primary reference and the retrieved neighbors. The output of this second generation step was used for evaluation. This condition tested whether an additional novelty-directed revision improved upon single-pass retrieval-augmented generation.

\subsubsection{Quantifying semantic space expansion}
A primary hallmark of scientific creativity is the ability to generate hypotheses that extend beyond the boundaries of existing knowledge. To assess whether ForeSCITE expands the training embedding landscape, we evaluate the global growth of the conclusion embedding space rather than focusing on granular, paper-level novelty. We utilized a density-based hypersphere coverage metric, \#Circles~\cite{xieHowMuchSpace2023}, to quantify semantic space coverage. Formally, let $Z_{emb}^{<t}$ represent the set of embeddings derived from the PubMed conclusions prior to the knowledge cut-off date t. Let $\hat{X}$ denote a set of conclusion-like hypotheses generated by a given model (either ForeSCITE or baselines) and let $\hat{Z}$ represent the corresponding set of embeddings generated via an embedding function, such that $\hat{Z}=E_{emb}(\hat{X}$). The coverage metric, \#Circles, calculates the volume (or area) in high-dimensional space covered by the union of hyperspheres centered at each embedding point, where the radius is determined by a density-based threshold. We define the Semantic Space Growth ($\Delta \text{Cov}$) as the marginal increase in embedding space volume provided by the generated hypotheses:

$$\Delta \text{Cov}=\#\mathrm{Circles}\left(Z_{emb}^{<t}\cup\hat{Z}\right)-\#\mathrm{Circles}\left(Z_{emb}^{<t}\right)$$

By comparing the $\Delta \text{Cov}$ of ForeSCITE against both the human-authored holdout conclusion embeddings ($Z_{emb}^{>t}$) and baselines, we can quantify the extent to which temporal extrapolation expands semantic coverage relative to the historical literature.

\subsubsection{Quantifying novelty and future validation ratio}
Beyond evaluating the expansion of the latent embedding space, we introduce an Entity Novelty Ratio and a Future Validation Ratio to quantify the novelty of generated entity pairings and their subsequent occurrence in the post-cutoff literature. Formally, let $\mathcal{G}^{<t}$ and $\mathcal{G}^{>t}$ denote the sets of co-occurring biomedical entity pairs extracted from the pre-cutoff training set and the post-cutoff holdout set, respectively, and let $\mathcal{F}=\mathcal{G}^{>t}\setminus\mathcal{G}^{<t}$ denote the novel post-cutoff co-occurrence set. Let $\hat{\mathcal{G}}$ denote the set of entity pairs extracted from generated hypotheses $\hat{X}$. The Entity Novelty Ratio measures the proportion of generated entity pairs absent from the training set:

$$\frac{\left|\hat{\mathcal{G}}-\mathcal{G}^{<t}\right|}{\left|\hat{\mathcal{G}}\right|}$$

A higher novelty ratio indicates the generation of entity pairs distinct from the training distribution. The Future Validation Ratio measures the fraction of all generated entity pairs that are absent from the training set and observed in the post-cutoff holdout set:

$$\frac{\left|\hat{\mathcal{G}}\cap\mathcal{F}\right|}{\left|\hat{\mathcal{G}}\right|}.$$

A higher validation ratio indicates a larger share of novel pairs subsequently reported in the post-cutoff conclusions. To construct the reference sets $\mathcal{G}^{<t}$ and $\mathcal{G}^{>t}$, we utilized pre-computed annotations from PubTator 3.0~\cite{weiPubTator30AIpowered2024}, sourced from BioCXML-formatted FTP dumps. To ensure the extracted entity pairs reflect scientific findings rather than background contexts, we strictly filter these relations to those whose entity spans reside within the parsed conclusion sections of the abstracts. As the generated hypotheses $\hat{X}$ do not exist within the PubTator 3.0 database, we processed them using the offline PubTator Tagger pipeline to ensure an equitable comparison.

\paragraph{Human-written reference for entity novelty.}
We applied the Entity Novelty Ratio definition to the human-written post-cutoff holdout conclusions, using $\mathcal{G}^{>t}$ as the evaluated entity-pair set. The resulting reference is $|\mathcal{F}|/|\mathcal{G}^{>t}|$, the fraction of post-cutoff pairs absent from the pre-cutoff training corpus. Across the 236,057 human-written post-cutoff conclusions, 161,625 of the 628,434 extracted pairs were novel, giving $161{,}625/628{,}434\approx25.72\%$. This value was calculated once over the full holdout corpus and used as a fixed empirical reference in both panels of Fig.~\ref{fig:coverage_symbolic_eval}c,d.

\paragraph{Chance baselines for future validation.}
We constructed two chance baselines using the pre-cutoff entity vocabulary $\mathcal{V}$, with $|\mathcal{V}|=317{,}547$. Eligible pairs consisted of two distinct entities in $\mathcal{V}$ and were required to be absent from $\mathcal{G}^{<t}$. Uniform sampling assigned equal probability to every eligible unseen pair. Degree-weighted sampling (the frequency-preserving baseline) drew endpoints with probabilities proportional to their pre-cutoff training degrees, using a configuration-model-inspired procedure, and rejected self-pairs and already-seen edges. Both procedures excluded duplicate pairs within each sampled set. The degree-weighted procedure favors historically well-connected entities without imposing exact preservation of the training degree sequence.

For each sampled set $\mathcal{S}$, we computed $|\mathcal{S}\cap\mathcal{F}|/|\mathcal{S}|$, using the same novel post-cutoff target set as for the generation methods. Because all sampled pairs were constrained to be unseen, their Entity Novelty Ratio was 1 by construction. These baselines therefore provide chance references for future overlap among unseen pairs within the pre-cutoff vocabulary, rather than natural-language hypothesis-generation baselines. Reported point estimates were analytic expectations under the respective pair-sampling models. To characterize finite-sample variability, we generated 1,000 sets of 200,000 unique unseen pairs for each baseline and obtained 99\% null sampling intervals from the resulting distributions of Future Validation Ratios. These intervals describe variation across random pair sets and are distinct from the hypothesis-level bootstrap distributions used for the generation methods.

\subsection{Model-based prioritization and expert assessment of CAR-T hypotheses}
\label{sec:car_t_expert_assessment}

We used ForeSCITE with $\alpha=0.25$ to generate 1,465 hypotheses concerning chimeric antigen receptor T (CAR-T) cell therapy. Each hypothesis was submitted to GPT-5 for novelty assessment using a fixed six-point rubric (0--5). The prompt defined novelty in terms of targets, mechanisms, tissue or immune interactions, and non-standard therapeutic combinations that were not broadly known or widely described in the CAR-T literature. Scores ranged from 0 (very low novelty) to 5 (very high novelty). GPT-5 was instructed to assess novelty alone, without evaluating correctness, feasibility or writing quality, and to return a JSON object containing an integer novelty score and a concise explanation. The request specified the model identifier \texttt{gpt-5}, medium reasoning effort, medium verbosity, and JSON-object response formatting. The prompts are provided in Appendix~\ref{app:car_t_novelty_prompt}.

Two immunology cell biologists reviewed all 240 hypotheses assigned novelty scores of 4 or 5, assessing each for novelty and biological plausibility. This review identified 37 hypotheses as satisfying both criteria. The proportion retained was calculated using the 240 reviewed hypotheses as the denominator; the 37 candidates were also reported as a proportion of all 1,465 generated hypotheses. Hypotheses with scores of 0--3 were not included in this screening review. We additionally present expert justifications for five selected hypotheses, discussing their biological rationale, novelty relative to earlier literature and subsequent published evidence where available. Of these five, three were drawn from the 37 hypotheses retained by expert review, one from the 240 reviewed hypotheses but outside the retained 37, and one from the full set of 1,465 generated CAR-T hypotheses. These qualitative examples were used to illustrate the interpretation of generated hypotheses.

\bibliographystyle{unsrt}
\bibliography{predmed}

\appendix
\setcounter{figure}{0}
\renewcommand{\thefigure}{\thesection\arabic{figure}}
\setcounter{table}{0}
\renewcommand{\thetable}{\thesection\arabic{table}}
\section{Pre-cutoff Reference Selection}

Table~\ref{tab:reference_selection} reports the publication-year distributions of the recency-based and impact-based pre-cutoff reference sets, each containing 236,057 conclusions. The recency-based set spans 2021--2023, whereas the impact-based set spans 2011--2023 and includes only conclusions from papers with RCR $>2$.

\begin{table}[!htbp]
    \centering
    \caption{Year-wise distribution of pre-cutoff reference conclusions under recency-based and impact-based selection. Impact-based references satisfy RCR $>2$; the listed counts total 236,057.}
    \label{tab:reference_selection}
    \begin{tabular}{lrr}
        \toprule
        Year & Recency-based selection & Impact-based selection \\
        \midrule
        2023 & 37,591 & 10,235 \\
        2022 & 141,502 & 27,605 \\
        2021 & 56,964 & 24,269 \\
        2020 & \textemdash & 21,715 \\
        2019 & \textemdash & 19,223 \\
        2018 & \textemdash & 17,431 \\
        2017 & \textemdash & 16,909 \\
        2016 & \textemdash & 16,700 \\
        2015 & \textemdash & 18,719 \\
        2014 & \textemdash & 18,121 \\
        2013 & \textemdash & 17,852 \\
        2012 & \textemdash & 16,995 \\
        2011 & \textemdash & 10,283 \\
        \midrule
        Total & 236,057 & 236,057 \\
        \bottomrule
    \end{tabular}
\end{table}
\section{HDBSCAN Minimum Cluster Size Selection}

\begin{figure}[!htbp]
    \centering
    \includegraphics[width=0.8\textwidth]{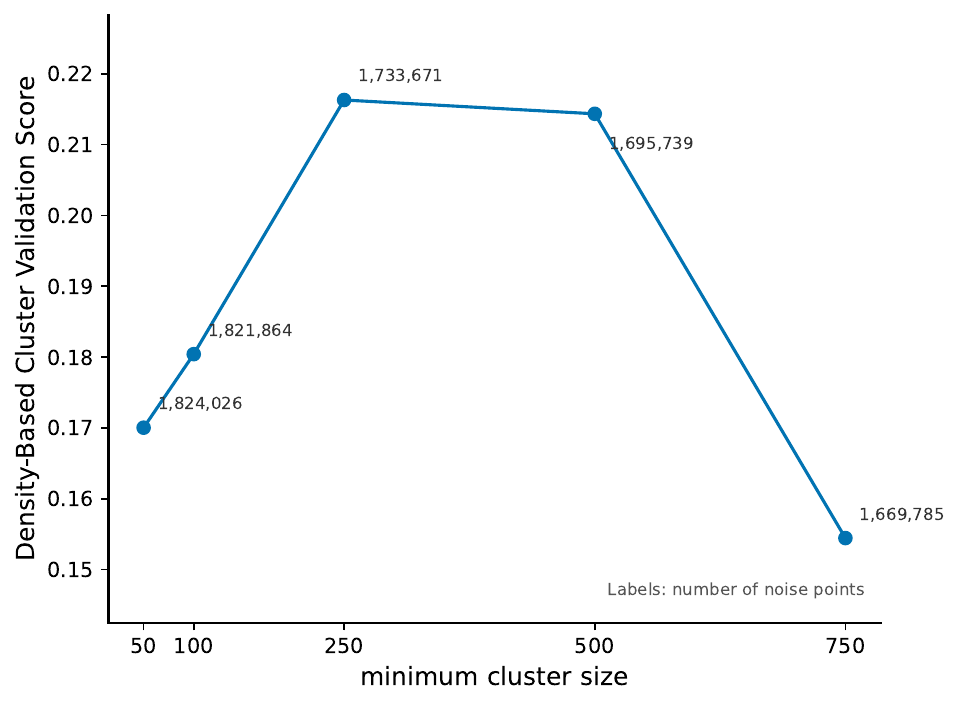}
    \caption{\textbf{Sensitivity analysis for the HDBSCAN minimum cluster size.} Density-based clustering validation (DBCV) score and the number of points assigned to noise across candidate minimum cluster sizes. A minimum cluster size of 500 retains a strong DBCV score while assigning fewer points to noise than a minimum cluster size of 250.}
    \label{fig:hdbscan_min_cluster_size}
\end{figure}

Figure~\ref{fig:hdbscan_min_cluster_size} compares the density-based clustering validation (DBCV) score and the number of points assigned to noise across candidate HDBSCAN minimum cluster sizes. We selected a minimum cluster size of 500 because it retains a strong DBCV score while assigning fewer points to noise than a minimum cluster size of 250.

\section{Temporal Step-Size Selection}

\begin{figure}[!htbp]
    \centering
    \begin{minipage}[t]{0.65\textwidth}
        \textbf{a}\par\vspace{0.5ex}
        \includegraphics[width=\linewidth]{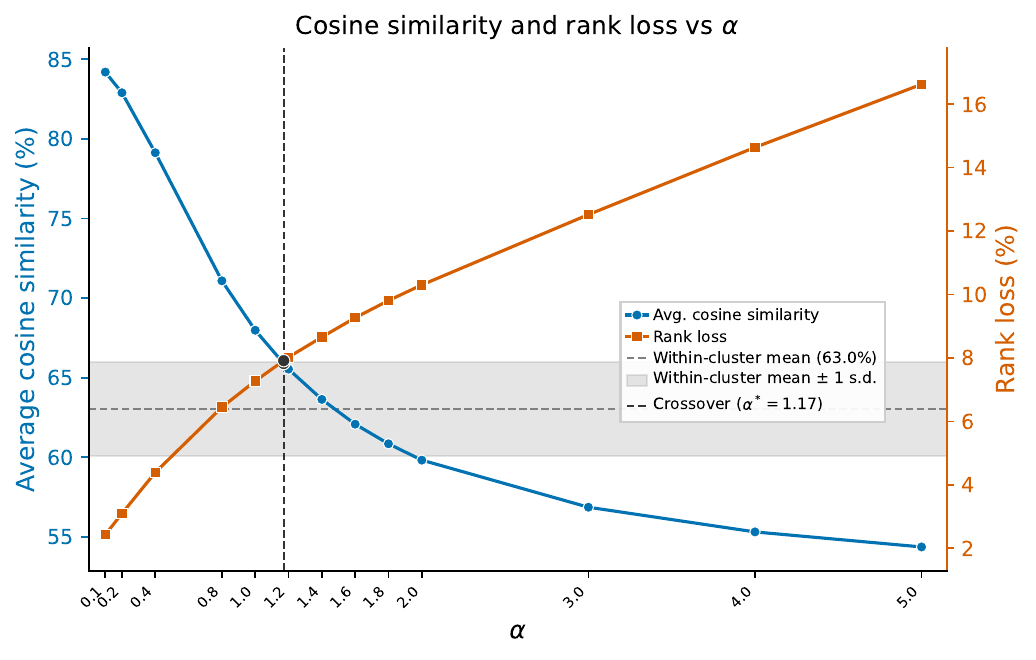}
    \end{minipage}
    \par\vspace{1.5ex}
    \begin{minipage}[t]{\textwidth}
        \textbf{b}\par\vspace{0.5ex}
        \centering
        \includegraphics[width=0.56\linewidth]{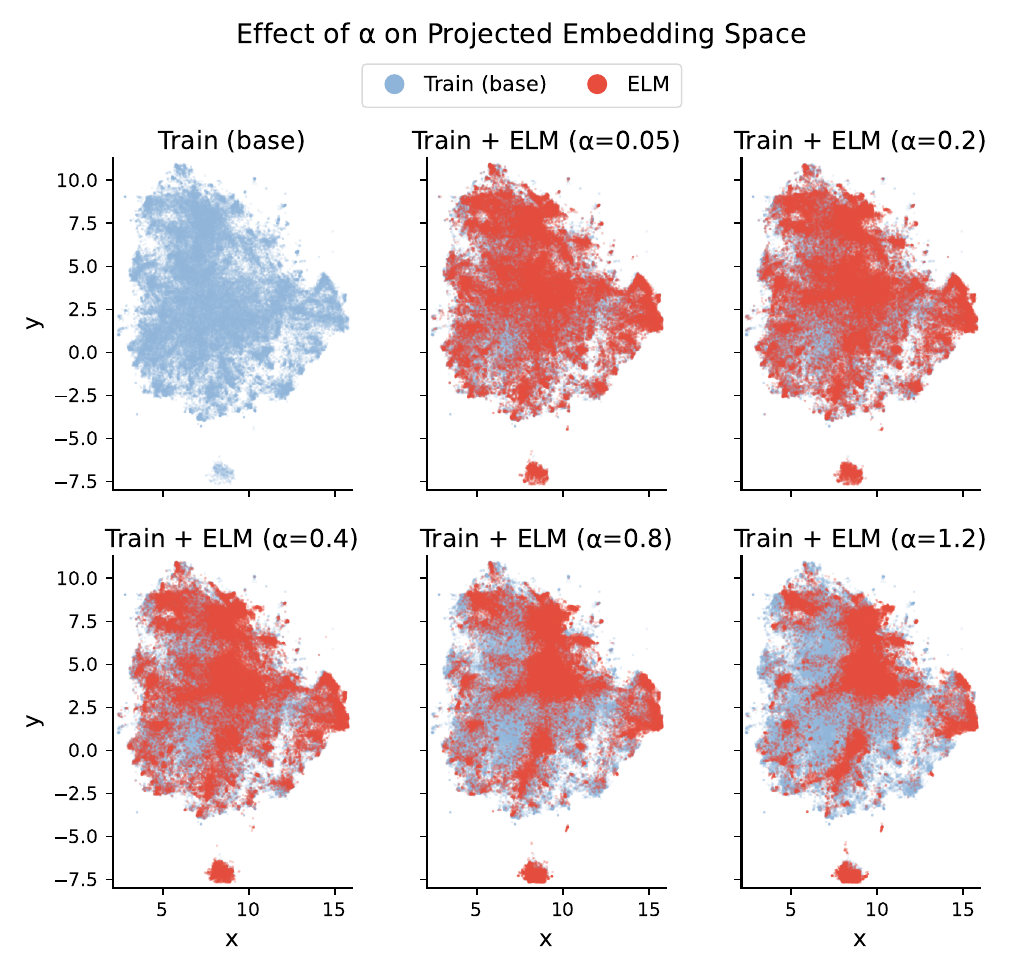}
    \end{minipage}
    \caption{\textbf{Extrapolation step size controls the tradeoff between semantic displacement and embedding diversity.} \textbf{a,} Average cosine similarity between 236,057 paired reference and ForeSCITE-generated conclusion embeddings (left axis) and effective-rank loss of the generated embedding matrix relative to the reference matrix (right axis) across temporal step sizes $\alpha$. The horizontal line and band show the mean within-cluster cosine similarity and $\pm1$ s.d.; the vertical line marks the estimated crossover at $\alpha^*=1.17$. Increasing $\alpha$ produces more distant embeddings but greater rank loss. \textbf{b,} Joint UMAP projection fitted to all training embeddings and ForeSCITE-generated embeddings at $\alpha\in\{0.05,0.2,0.4,0.8,1.2\}$. Generated embeddings broadly follow the training manifold at small $\alpha$ but become progressively concentrated as $\alpha$ increases, illustrating space collapse.}
    \label{fig:cav_alpha_analysis}
\end{figure}

Figure~\ref{fig:cav_alpha_analysis} summarizes the effect of the temporal step size $\alpha$ on semantic displacement and embedding diversity: panel~a shows paired reference--generation cosine similarity and effective-rank loss, and panel~b shows the distributions of training and generated embeddings in a shared UMAP projection. These diagnostics motivated the operating regime $\alpha<1.0$ for downstream hypothesis generation, with larger values included as sensitivity analyses of semantic drift and space collapse.

\section{Baseline Prompt Templates}
\label{app:baseline_prompts}

In the templates below, \verb|{reference}| denotes the evaluation conclusion and \verb|{top_k_results}| denotes a newline-separated list of the five retrieved nearest-neighbor conclusions.

\subsection{Direct Prompt}

\begin{verbatim}
Reference conclusion: {reference}

Write a possible future conclusion that builds on this conclusion.
Output only the conclusion text.
\end{verbatim}

\subsection{Temporal Prompt}

\begin{verbatim}
Reference conclusion: {reference}

Write a hypothetical future conclusion that might emerge later from
novel hypotheses. Output only the new conclusion, nothing else.
\end{verbatim}

\subsection{Novel Prompt}

\begin{verbatim}
Reference conclusion: {reference}

Generate a future conclusion based on a novel hypothesis building on
this. Do not repeat or paraphrase the reference. Output only the new
conclusion.
\end{verbatim}

\subsection{Novel RAG and RAG + Iter., Turn 1}

The same first-turn template was used for Novel RAG and RAG + Iter.

\begin{verbatim}
Primary Reference Conclusion: {reference}

Similar Existing Conclusions:
{top_k_results}

Generate a future conclusion based on a novel hypothesis building on
these. Do not repeat or paraphrase the references. Output only the new
conclusion.
\end{verbatim}

\subsection{RAG + Iter., Turn 2}

\begin{verbatim}
Your conclusion has similarities with existing conclusions as provided
at the beginning of this conversation. Make sure the conclusion you
suggest is significantly different from the reference and existing
conclusions. Let's give it another try.
\end{verbatim}

\section{CAR-T Hypothesis Novelty Assessment}
\label{app:car_t_novelty_prompt}

GPT-5 evaluated the novelty of each ForeSCITE-generated CAR-T hypothesis using the following developer-role instructions and user-role template. The placeholder \verb|{testing_conclusion}| was replaced with the hypothesis being evaluated. Request parameters are listed separately from the prompt text.

\subsection{Developer-role prompt}

\begingroup
\lstset{literate={“}{{\textquotedblleft}}1 {”}{{\textquotedblright}}1}
\begin{promptbox}
- You are an expert in CAR-T cell therapy, immuno-oncology, and translational cancer biology.
- Your task is to evaluate the scientific novelty of hypothesis-like statements.
- Novelty is defined as how much the main claim introduces new targets, new mechanisms, new tissue/immune interactions, or non-standard therapeutic combinations that are *not already broadly known or widely described in CAR-T therapy literature*.
- Use the following novelty scoring rubric:

0 = Very low novelty (already well-known, common review-language, generic statements like “CAR-T is promising”)
1 = Low novelty (known strategy or combination, widely reported in standard CAR-T clinical review literature)
2 = Moderate-low novelty (some distinctive angle, but still consistent with widely explored CAR-T strategies)
3 = Moderate novelty (less commonly discussed, applies known CAR-T concepts to a more specific or less-studied cancer subtype or target)
4 = High novelty (proposes a less-studied target, mechanism, or combination with a plausible biological basis but not widely covered in existing CAR-T literature)
5 = Very high novelty (introduces a new or unusual mechanistic insight, target, or therapeutic strategy not previously reported in CAR-T research to date)

- DO NOT evaluate correctness, feasibility, or writing quality. Only evaluate *novelty relative to existing research directions that are widely explored*.
- Produce output in exactly this JSON format, with no additional commentary:
{
    "novelty_score": <0-5>,
    "reasoning": "<one or two concise sentences explaining the score>"
}
\end{promptbox}
\endgroup

\subsection{User-role prompt template}

\begin{promptbox}
Evaluate the following hypothesis for novelty:

{testing_conclusion}
\end{promptbox}

\end{document}